\documentclass{article} 
\usepackage[final]{colm2026_conference}

\usepackage{microtype}
\usepackage{hyperref}
\usepackage{url}
\usepackage{booktabs}
\usepackage{amsmath}
\usepackage{graphicx}
\usepackage{subcaption}
\usepackage{array}
\usepackage{multirow}
\usepackage[most]{tcolorbox}
\usepackage{xcolor}
\DeclareMathOperator*{\argmax}{argmax}
\DeclareMathOperator*{\argmin}{argmin}
\usepackage[export]{adjustbox}
\usepackage{placeins}

\usepackage{lineno}

\definecolor{darkblue}{rgb}{0, 0, 0.5}
\hypersetup{colorlinks=true, citecolor=darkblue, linkcolor=darkblue, urlcolor=darkblue}

\title{Minimal, Local, Causal Explanations for Jailbreak Success in Large Language Models}

\author{Shubham Kumar \& Narendra Ahuja  \\
Department of Electrical Engineering\\
University of Illinois at Urbana-Champaign\\
\texttt{\{sk138,n-ahuja\}@illinois.edu}
}

\begin{document}

\ifcolmsubmission
\linenumbers
\fi

\maketitle

\begin{abstract}
Safety trained large language models (LLMs) can often be induced to answer harmful requests through \textit{jailbreak prompts}. Because we lack a robust understanding of why LLMs are susceptible to jailbreaks, future frontier models operating more autonomously in higher-stakes settings may similarly be vulnerable to such attacks. Prior work has studied jailbreak success by examining the model's intermediate representations, identifying directions in this space that causally encode concepts like harmfulness and refusal. Then, they \textit{globally} explain all jailbreak attacks as attempting to reduce or strengthen these concepts (e.g., reduce harmfulness). However, different jailbreak strategies may succeed by strengthening or suppressing different intermediate concepts, and the same jailbreak strategy may not work for different harmful request categories (e.g., violence vs. cyberattack); thus, we seek to give a \textit{local} explanation---i.e., why did this specific jailbreak succeed? To address this gap, we introduce \textbf{LOCA}, a method that gives Local, CAusal explanations of jailbreak success by identifying a minimal set of interpretable, intermediate representation changes that causally induce model refusal on an otherwise successful jailbreak request. We evaluate LOCA on harmful original–jailbreak pairs from a large jailbreak benchmark across Gemma, Llama, and Qwen chat models, comparing against prior methods adapted to this setting. LOCA can successfully induce refusal by making, on average, six interpretable changes; prior work routinely fails to achieve refusal even after 20 changes. LOCA is a step toward mechanistic, local explanations of jailbreak success in LLMs. Code is publicly available: \url{https://github.com/skumar-ml/loca-jailbreaks}
\end{abstract}

\section{Introduction}
\label{intro}

Large language models (LLMs) have continued to grow in their capabilities, and with the rise of agentic AI, they are being used more autonomously for higher-stakes settings \citep{Bommasani2021OnTO, Wang2023ASO}. Because of their potential for misuse, LLMs deployed to the public typically undergo \textit{alignment} fine-tuning in order to learn to refuse harmful (e.g., safety policy violating) requests while providing helpful answers on harmless requests \citep{Bai2022TrainingAH}; however, user can craft \textit{jailbreak} attacks, which are harmful requests that bypass LLM refusal, eliciting a harmful response \citep{chu-etal-2025-jailbreakradar}. 

To understand LLM refusal behavior, a growing body of mechanistic interpretability work has identified directions in LLM intermediate representation space that causally encode concepts like harmfulness and refusal \citep{geometry_of_refusal, arditi2024refusal, Ball2024UnderstandingJS}. Then, these concepts are used to produce a \textit{global}---across all inputs---explanation for refusal behavior. For example, some prior works find that jailbreaks succeed by reducing the degree of harmfulness in intermediate representations \citep{arditi2024refusal, Ball2024UnderstandingJS}. However, different jailbreak strategies may succeed by strengthening or suppressing different intermediate concepts, and the same jailbreak strategy may not work for different request categories (e.g., violence vs. cyberattack). We believe that jailbreak success may be more nuanced than what global explanations can capture, motivating the need for \textit{local}, sample-specific explanations. Furthermore, we want our explanation to be \textit{causal}; that is, our explanation should isolate aspects of the intermediate representation that, when intervened on, induce refusal behavior on an otherwise successful jailbreak request. Finally, we desire a \textit{minimal} or parsimonious explanation to ensure we identify the most important causal interventions; this results in an explanation more conducive to human understanding and interpretation \citep{2010TheMM, Miller1956TheMN}. We propose \textbf{LOCA} as a method to find minimal, LOcal, CAusal explanations of jailbreak success. Our contributions are:

\begin{enumerate}
    \item LOCA can induce refusal on an otherwise successful jailbreak request \textbf{by making (on average) six interventions} on intermediate representations from a Llama chat model. Other methods are generally unable to induce refusal even after 20 interventions.

    \item \textbf{An ablation study validates that LOCA's algorithmic novelties} drive improvement, explaining why LOCA outperforms prior methods.


    \item A localization analysis finds that changes to \textbf{instruction (i.e., user request) tokens are the most causally important for inducing refusal in earlier layers}. LOCA induces refusals in later layers by relying almost exclusively on punctuation and post-instruction (i.e., chat template) tokens.
        
    \item \textbf{A case study} on using LOCA to explain an instance of jailbreak success.    
    
\end{enumerate}

\section{Related works}
\label{related}

\textbf{Finding interpretable linear directions, or concepts, in LLMs:}
The linear representation hypothesis posits that meaningful \textit{concepts} are encoded as \textit{linear directions} in a model's representation space \citep{park2023thelrh, Elhage2022ToyMO}. For example, prior work has found directions corresponding to truth \citep{Zou2023RepresentationEA} and knowledge awareness \citep{ferrando2025doIKnow}. These directions can be found in a supervised manner by training \textit{probes} on intermediate representations, or \textit{activations}, to separate between positive and negative examples of a concept \citep{Cunningham2026ConstitutionalCE}. Aside from probing, \textit{sparse autoencoders} (SAEs) have emerged as a powerful, unsupervised tool to surface a large set of disentangled interpretable concepts \citep{bricken2023monosemanticity, Cunningham2023SAE}. 

To causally interpret the concept encoded by a given direction (either from a probe or an SAE), prior works analyzes the impact of changing activations along the specified direction during the LLM's forward pass. These changes, or \textit{interventions}, typically happen by either activation steering or activation patching. \textit{Activation steering} adds (or subtracts) the scaled direction to activations at many or all token positions \citep{Turner2023SteeringLM}. It is critical to scale the operation properly; a scale too small may result in no change in the output, and a scale too large may push the model off-manifold, resulting in nonsensical outputs. \textit{Activation patching} makes interventions by replacing activations during a forward pass on the target prompt with reference activations from a different, reference prompt \citep{meng2022locating, Heimersheim2024ActPatchTutorial}. Because we have reference activations, it is easier to make targeted and varied changes per-token. Activation patching commonly uses a reference prompt from a closely matched template to align activations across corresponding token positions; otherwise, defining the token correspondence becomes less clear.

\textbf{Jailbreak attacks and defenses:} Jailbreak attacks take many forms. Some use hand-crafted strategies, such as role-play; others obfuscate harmful intent through encoding, translation, or low-resource languages; automated attacks optimize adversarial suffixes or prompt rewrites to maximize compliance \citep{zou2023universal, Chao2023JailbreakingBB, liu2024autodan}. More recent attacks also exploit multi-turn interaction, decomposition, or indirect prompting, making jailbreaks difficult to characterize as a single prompt pattern \citep{Russinovich2024GreatNW, srivastav-zhang-2025-safe}. Correspondingly, jailbreak defenses span several broad categories. Input-level heuristics can detect or disrupt jailbreak prompts before generation \citep{Jain2023BaselineDF}. \citet{Inan2023LlamaGL} propose a separate guardrail model around the target LLM, either to classify unsafe inputs/outputs or to revise unsafe responses. The LLM itself can also be fine-tuned to be more robust to jailbreak attempts \citep{Bai2022TrainingAH}.

Most closely related to our setting are inference-time, activation-based defenses, which intervene directly on the model's internal representations \citep{Zhang2025JBShieldDL, zhao-etal-2025-adasteer, Zhu2026PrincipledSV, Assogba2026SparseAA}. For example, JBShield \citep{Zhang2025JBShieldDL} extracts toxic and jailbreak concepts from hidden states, which inform a global steering shift for mitigating detected jailbreak attempts. LOCA differs from such works by shifting the jailbreak mechanism from global (for all inputs) to local (for specific inputs), enabling a token-level analysis of jailbreak behavior.

In parallel, mechanistic work investigates which internal components implement safety behavior: for instance, \citet{zhou2025on} identify safety-critical attention heads whose ablation substantially weakens refusal behavior, localizing safety parameters further than prior residual stream analyses. These works, in contrast to LOCA, focus on localization rather than explanation of jailbreak behavior.

\textbf{Global understanding of jailbreaks in an LLM's representation space:}
Much of the literature has concentrated on understanding jailbreaks by looking for global explanations in the intermediate representation space of LLMs. Some works suggest that refusal is mediated by harmfulness; thus, jailbreaks succeed by suppressing input token projections onto these harmfulness directions \citep{arditi2024refusal, Ball2024UnderstandingJS, lin-etal-2024-towards-understanding}. Complementary work by \citet{geometry_of_refusal} finds that a gradient-based optimization can yield a refusal \textit{subspace} that is more suitable for causally controlling refusal (e.g., by steering the model). However, other work offers a more nuanced perspective. A study from \citet{Zou2023RepresentationEA} finds that in 90\% of successful jailbreaks, the model accurately represents the request as harmful, yet does not refuse. Furthermore, \citet{bau_refusal_and_harm_seperate} finds causal evidence that LLMs represent harmfulness and refusal separately: refusal can be bypassed even if the model knows the question is harmful. This suggests that while refusal can be represented in a low-dimensional subspace, refusal itself is determined by multiple internal concepts, not just harmfulness. Furthermore, these concepts may not be global---they may vary greatly among different jailbreak examples. \citet{bau_refusal_and_harm_seperate, kirch-what-features-jailbreak} find that different jailbreak strategies try to bypass refusal through different mechanisms, and different request risk categories may rely on different concepts to induce refusal.

There have been two previous attempts at characterizing refusal-causing concepts by looking at early (or upstream) layer representations. \citet{lee2025upstreamrefusal} find causal, early-layer SAE directions (or vectors) that, when used for steering, increase downstream request token embedding projections onto the refusal direction (found from \citet{arditi2024refusal}). On three hand-selected prompts, this method surfaced upstream causal steering vectors that induced refusal.
\citet{yeo-2025-understanding-refusal-saes} find upstream SAE vectors causal to refusal with a two-step process: (1) take the top-$M$ SAE directions that have the highest cosine similarity with the refusal direction and (ii) take the top-$K$ (where $K<M$) SAE vectors that lead to the largest change on output token probabilities after activation patching from a harmful, non-refused instruction to a harmful, refused instruction. They validate these SAE vectors by steering, showing that the causal vectors can either induce or bypass refusal.

Both works successfully find upstream vectors causal to refusal, but these concepts are not used to \textit{locally explain why} a jailbreak succeeded. Additionally, both works are limited by two weaknesses. First, they use first-order approximations that \textbf{are averaged across tokens}; consequently, they struggle to localize interventions to specific tokens, so they make interventions along all tokens. Second, the top-$K$ vectors are selected \textbf{in one-shot}, which ignores interaction effects introduced when steering or patching with the selected vectors. LOCA addresses both limitations, allowing us to explain jailbreak success in terms of the minimal, causal change needed to recover the original refusal response.

\section{LOCA: LOcal, CAusal explanations of jailbreak success}
\label{method}
Consider the following setting. The user wants an LLM to answer the original, target prompt $x_o$, but the LLM has gone through alignment and refuses answering $x_o$. The user creates a jailbreak prompt $x_j$ to bypass the alignment and elicit an answer to the original $x_o$. To explain why $x_j$ succeeded, we are interested in finding a minimal change to $x_j$ in the representation space that causes $x_j$ to fail (i.e., induce a similar refusal response to $x_o$).

To make the change, we can either perform activation steering or activation patching. \cite{lee2025upstreamrefusal} performed steering, which we believe to be suboptimal because (1) it may result in off-manifold representations and (2) there is not a principled way to select the tokens to steer. Thus, we opt to perform activation patching, which allows for token-specific, within-distribution changes. To enable activation patching between two structurally different prompts, LOCA first matches tokens from $x_j$ to $x_o$. Then, LOCA iteratively ranks the per-token, per-concept changes using a first-order approximation and applies activation patching to the intermediate token representations, or token \textit{embeddings}, of $x_j$.

\subsection{Preliminaries}
\textbf{Transformers.} Decoder-only transformers \citep{vaswani2017attention} tokenize a prompt $x$ into a sequence of input tokens $T = (\mathbf{t}_1, \mathbf{t}_2, ..., \mathbf{t}_N)$, where the number of tokens $N$ depends on $x$. Each token $\mathbf{t}_i$ is embedded to $\mathbf{h}_i^{(1)} \in \mathbb{R}^D$. The transformation from each transformer layer $l$ is written back to the residual stream, updating the intermediate token embedding as $\mathbf{h}_i^{(l)}$. After the last layer, the model predicts the next output token as logits $\mathbf{y} \in \mathbb{R}^{|V|}$ over the vocabulary $V$, which is converted to probabilities $\mathbf{p}$ via softmax. Similar to prior studies \citep{arditi2024refusal, bau_refusal_and_harm_seperate, lee2025upstreamrefusal, yeo-2025-understanding-refusal-saes}, we analyze intermediate representations from the residual stream.

\textbf{Activation Patching.} Given two sequences $T_{reference}$ and $T_{target}$, activation patching evaluates the effect of overwriting (or \textit{patching}) selected intermediate activations from $T_{target}$ with corresponding activations from $T_{reference}$. The \textit{patching effect} measures how much the LLM output changed---specifically, how closely the output from the patched $T_{target}$ recovers $\mathbf{y}_{reference}$. One typically activation patches entire embeddings, but one can also activation patch after projecting the embeddings onto a subspace.

\textbf{Sparse Autoencoders.} Sparse autoencoders (SAEs) are a family of autoencoders that map an input activation $\mathbf{x} \in \mathbb{R}^D$ to a feature vector $\mathbf{f} \in \mathbb{R}^M$ and then reconstruct the input with a linear decoder $W_d$. SAEs typically take the form of
\begin{equation}
    \mathbf{f} = \phi(W_e \mathbf{x} + \mathbf{b}_e), \quad \hat{\mathbf{x}} = W_d \mathbf{f} + \mathbf{b}_d
\end{equation}
where $\phi$ is a non-linear function. SAEs are trained to minimize reconstruction error while satisfying some sparsity objective on $\mathbf{f}$. Each row $\mathbf{v_i} \in \mathbb{R}^D$ of $W_d$ is a \textit{concept vector}, denoting an interpretable direction in the original representation space. LOCA will activation patch after projecting embeddings onto these concept vectors.

\begin{figure*}[b]
    \centering

    \begin{subfigure}[t]{0.32\textwidth}
        \centering
        \includegraphics[width=\linewidth]{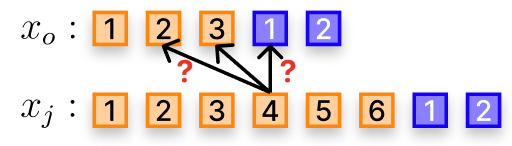}
        \caption{Token matching issue}
        \label{fig:loca-token-matching-a}
    \end{subfigure}
    \hfill
    \begin{subfigure}[t]{0.32\textwidth}
        \centering
        \includegraphics[width=\linewidth]{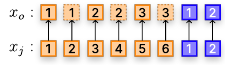}
        \caption{Re-sample \textcolor{orange}{$T_{\text{inst}}$}, then match}
        \label{fig:loca-token-matching-b}
    \end{subfigure}
    \hfill
    \begin{subfigure}[t]{0.32\textwidth}
        \centering
        \includegraphics[width=\linewidth]{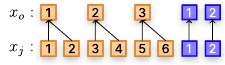}
        \caption{Final matching scheme}
        \label{fig:loca-token-matching-c}
    \end{subfigure}

    \caption{\textbf{LOCA token matching:} The jailbreak prompt $x_j$ and original (reference) prompt $x_o$ can be of any format, making naive token matching ill-specified (Fig. \ref{fig:loca-token-matching-a}). LOCA handles this by resampling (in this case, upsampling) $x_o$'s \textcolor{orange}{$T_{\text{inst}}$} tokens to match the length of $x_j$'s \textcolor{orange}{$T_{\text{inst}}$} tokens (Fig. \ref{fig:loca-token-matching-b}). Then, \textcolor{orange}{$T_{\text{inst}}$} and \textcolor{blue}{$T_{\text{post-inst}}$} tokens can be matched one-to-one (Fig. \ref{fig:loca-token-matching-c}).}
    \label{fig:loca-token-matching}
\end{figure*}

\subsection{Token matching}
Activation patching is conventionally used when prompts $x_{reference}$ and $x_{target}$ follow the same structure, since the correspondence between \textit{reference} tokens and \textit{target} tokens is straightforward. This is problematic in our setting, where we are trying to patch between the original (reference) prompt $x_o$ and a jailbreak (target) prompt $x_j$, because the jailbreak can be of any format and length as long as it is successful (Fig. \ref{fig:loca-token-matching-a}). LOCA handles this with a token matching scheme (shown in Fig. \ref{fig:loca-token-matching-b}, \ref{fig:loca-token-matching-c}). A prompt $x$ to an instruct (or chat) LLM will follow a chat template, resulting in the following three components: (1) system tokens (\textcolor{gray}{$T_{\text{sys}}$}), (2) instruction tokens (\textcolor{orange}{$T_{\text{inst}}$}), and (3) post-instruction tokens (\textcolor{blue}{$T_{\text{post-inst}}$}):
\begin{enumerate}
    \item \textcolor{gray}{$T_{\text{sys}}$} tokens (and their embeddings) are the same for any $x$ in causal, decoder-only transformers, so we safely ignore them.
    
    \item \textcolor{orange}{$T_{\text{inst}}$} tokens from $x_o$ and $x_j$ have variable length. $x_o$'s \textcolor{orange}{$T_{\text{inst}}$} are upsampled or downsampled to match the length of $x_j$'s \textcolor{orange}{$T_{\text{inst}}$}, allowing us to match one-to-one. When upsampling, we simply repeat the upsampled token embedding (no interpolation). When downsampling, we skip over embeddings. 

    \item \textcolor{blue}{$T_{\text{post-inst}}$} tokens are the same for any prompt $x$, though naturally their embeddings are different. Embeddings at these positions are matched one-to-one.
\end{enumerate}

While arbitrary, we find that this token matching scheme works well in practice\footnote{We experiment with other token matching schemes in Appendix \ref{sec:appendix-sentivity-token-matching}. We do not find significant differences in the results, suggesting that the token matching scheme does not matter significantly.}. We denote this scheme as function $\mathcal{M}:\mathbb{R} \to \mathbb{R}$, which takes in a \textit{target} token index and returns the matched \textit{reference} token index.

\subsection{Patching effect measure}
\label{sec:patching-effect-measure}
Denote the patched jailbreak embedding sequence at layer $l$ as $\tilde{H}^{(l)}_j$. Ideally, our patching effect measures the full difference between the entire generated response to $H^{(l)}_o$ and $\tilde{H}^{(l)}_j$. However, it is computationally expensive to generate the full response to $\tilde{H}^{(l)}_j$ for each patching operation. Thus, it is common to define a measure in terms of the first output token prediction $\mathbf{p}$.\footnote{Using the first-token is a cheap proxy that empirically works well. Some empirical evidence is given in \citet{arditi2024refusal}; we provide our own analysis in Appendix \ref{sec:appendix-justifying-first-token-proxy}.} To avoid focusing on specific vocabulary dimensions, we choose $\mathcal{L} = KL(\mathbf{p}_o || \mathbf{\tilde{p}}_j)$, where $KL(a||b)$ denotes the KL divergence between probability distribution $b$ and a reference distribution $a$. From here on, we drop the superscript $l$.

\subsection{Token-specific first-order approximation to patching along SAE directions}
It is intractable to compute the first-token prediction $
\mathbf{\tilde{p}}_j$ for all possible activation patches. Instead, we use the first-order approximation of the patching effect when patching the $i$'th jailbreak embedding in layer $l$ along any direction $\mathbf{v}$. Letting $h_{o,i}$ and $h_{o,j}$ respectively represent the $i$'th original and jailbreak embedding, our approximation is:
\begin{equation}
\label{eq:approx}
d(i, \mathbf{v}; \mathbf{p}_o, \mathbf{p}_j, l)
=
\underbrace{\left[\nabla_{\mathbf{h}_{j,i}} KL(\mathbf{p}_o \,\|\, \mathbf{p}_j)^T \mathbf{v}\right]}_{\text{directional derivative}}
\underbrace{(\mathbf{h}_{o,\mathcal{M}(i)} - \mathbf{h}_{j,i})^T \mathbf{v}}_{\text{magnitude term}}
\end{equation}
To develop an explanation, it is critical that patching occurs along interpretable directions $\mathbf{v}$. Thus, we patch along the concept vectors in the corresponding layer's SAE decoder $W_{d}$, which can be interpreted with interfaces such as Neuronpedia \citep{neuronpedia}. Notice how our approximation\footnote{Appendix \ref{sec:appendix-first-order-approx-error} empirically analyzes the accuracy of this first-order approximation on \textsc{Llama}, finding low approximation error. Any significant error is localized to the initial layers.} is token-specific (denoted by $i$), as opposed to prior work.

\begin{figure*}[t]
    \centering
    \includegraphics[width=\linewidth]{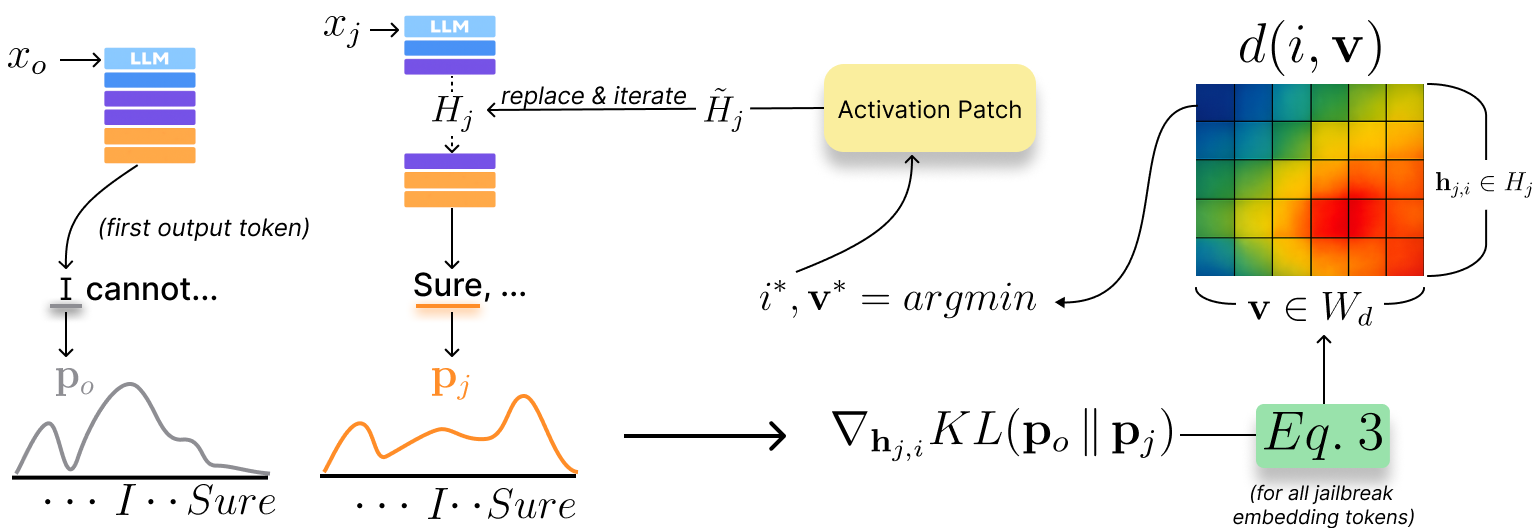}
    \caption{\textbf{LOCA algorithm:} LOCA computes the first output token probabilities $\mathbf{p}_o$ and $\mathbf{p}_j$ of the refused original $x_o$ and successful jailbreak $x_j$ prompt. Then, it selects a jailbreak embedding $\mathbf{h}_{j,i^*}$ to activation patch along direction $\mathbf{v}^*$ to minimize the KL divergence of the two distributions. The minimization is done over a first-order approximation (Eq. \ref{eq:iter-approx}), and $\mathbf{v}$ is selected from a SAE decoder $W_d$. This procedure is applied iteratively to induce refusal.}
    \label{fig:loca-overview}
\end{figure*}

\subsection{Operationalizing LOCA as an iterative algorithm}
To give a minimal, local, causal explanation of jailbreak success, LOCA uses an iterative algorithm (Fig. \ref{fig:loca-overview}). We denote the iterative version of Eq. \ref{eq:approx} at iteration $\alpha$ as:
\begin{equation}
\label{eq:iter-approx}
    d^{(\alpha)}(i, \mathbf{v})  = [\nabla_{\mathbf{h}^{(\alpha)}_{j, i}} KL(\mathbf{p}_o || \mathbf{p}^{(\alpha)}_j)^T\mathbf{v}](\mathbf{h}_{o, \mathcal{M}(i)} - \mathbf{h}^{(\alpha)}_{j, i})^T\mathbf{v}
\end{equation}
Then, the iterative algorithm proceeds as follows:
\begin{enumerate}
    \item Set $\alpha=0$. Initialize $\mathbf{p}^{(0)}_j=\mathbf{p}_j$, $\mathbf{h}^{(0)}_{j, i}=\mathbf{h}_{j, i} \; \forall \; i=1...N_j$, where $N_j$ is the number of jailbreak tokens.
    \item Find the minimizer:
    \[
    i^{(\alpha)^*}, \mathbf{v}^{(\alpha)^*} = \argmin_{i=1...N_j, \mathbf{v} \in W_{d} } d^{(\alpha)}(i, \mathbf{v})
    \]
    \item Activation patch $\mathbf{h}^{(\alpha+1)}_{j, i^{(\alpha)^*}}$ with the matched $\mathbf{h}_{o, \mathcal{M}(i^{(\alpha)^*})}$, but only along direction $\mathbf{v}^{(\alpha)^*}$. The remaining jailbreak embeddings are unchanged from the previous iteration. This results in embedding sequence $H_j^{(\alpha+1)}$, which is used to     complete the forward pass to compute $\mathbf{p}^{(\alpha+1)}_j$.
    %
    \item Set $\alpha:=\alpha + 1$ and repeat Step 2 and 3. Terminate upon meeting a stopping criterion.
\end{enumerate}
Appendix \ref{sec:appendix-loca-patching} contains more details on the patching operation. By recalculating the first-order approximation in every iteration (Eq. \ref{eq:iter-approx}), our approximations are conditioned on prior activation patching steps, improving over prior work.

\section{Evaluating \& Analyzing LOCA}
First, we compare LOCA to prior work \citep{lee2025upstreamrefusal, yeo-2025-understanding-refusal-saes}, showing that LOCA induces refusal at a significantly higher rate and requires significantly fewer patches than other methods. LOCA's effectiveness is validated with an ablation study, and a localization analysis gives insights into where refusal is determined in earlier layers.

\textbf{Models.} We study the Gemma-2-2B-IT (\textsc{Gemma-2}), Gemma-3-27B-IT (\textsc{Gemma-3}), Qwen-3-8B (\textsc{Qwen}), and Llama-3.1-8B-Instruct (\textsc{Llama}) chat models \citep{Riviere2024Gemma2I, Dubey2024TheL3, gemmateam2025gemma3technicalreport, yang2025qwen3technicalreport}. These models are safety aligned and instruction-tuned (IT), and they generally refuse canonical harmful requests \citep{arditi2024refusal}. 

\textbf{SAEs.} For \textsc{Gemma-2} and \textsc{Gemma-3}, we use the corresponding open-source GemmaScope SAEs, which are trained on text from the same distribution as \textsc{Gemma}'s pretraining data (web documents, code, and scientific articles) \citep{lieberum-etal-2024-gemmascope, gemmascope2}. Although the GemmaScope SAEs are trained on activations from the base model, prior work shows that they transfer reasonably well to IT-models without additional finetuning \citep{, yeo-2025-understanding-refusal-saes, lieberum-etal-2024-gemmascope, kissane2024saetransfer}. For \textsc{Llama}, we use the open-source SAEs from \citet{ardidi2025misalignedpersonas}, which are trained on a mix of pre-training, chat, and misaligned data; this makes the SAE more likely to surface jailbreak-relevant concepts. For \textsc{Qwen}, we use the QwenScope SAEs \citep{qwenscope}.

\textbf{Baselines. } We use the prior methods in \citet{lee2025upstreamrefusal} and \citet{yeo-2025-understanding-refusal-saes} as the comparative baselines. Implementation details and efforts taken to adapt these methods to our setting are in Appendix \ref{sec:appendix-baseline-methods}.

\textbf{Datasets.} We use the WhatFeatures dataset \citep{kirch-what-features-jailbreak}, a collection of 10,800 jailbreak attacks generated from 35 jailbreak methods. For each model, we generate and save responses to each jailbreak attack, using greedy decoding with a generation length of 512 tokens. Jailbreak success is labeled with the HarmBench autograder \citep{mazeika2024harmbench}. We randomly split this dataset into train, validation, and test sets using a 70/10/20 split. The train and validation set is used only by \citet{lee2025upstreamrefusal} to find the refusal direction.

\begin{figure}[t]
    \centering

    \begin{subfigure}[t]{0.45\textwidth}
        \centering
        \includegraphics[width=\linewidth]{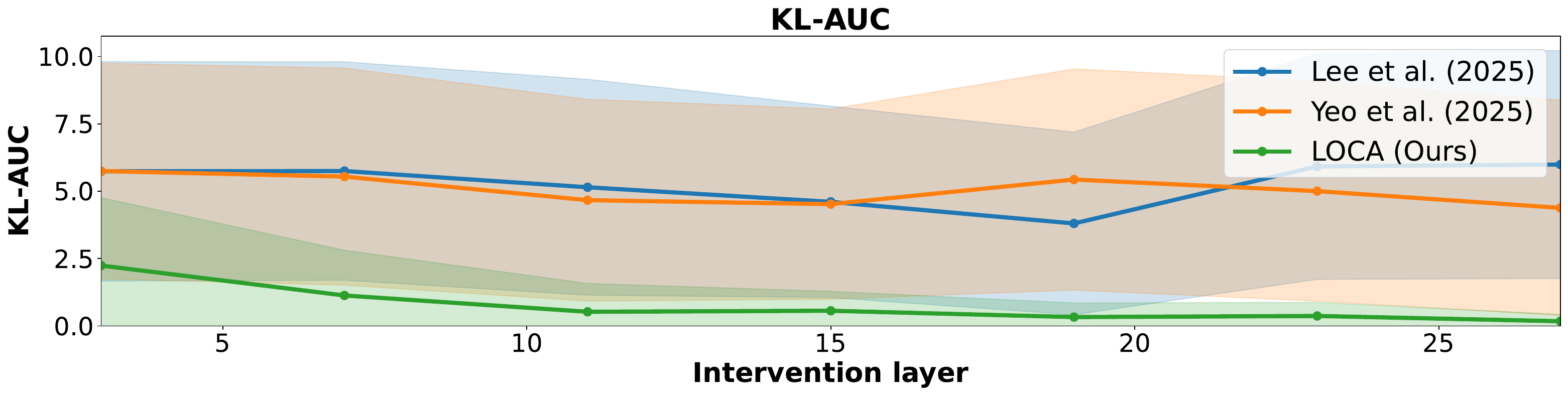}        
        
        \includegraphics[width=\linewidth]{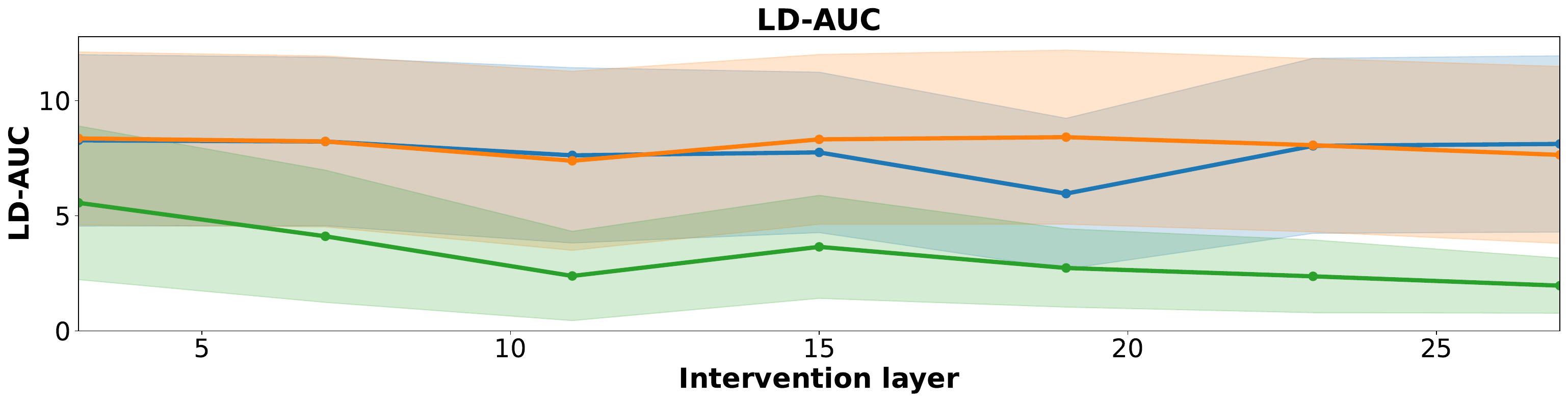}        
        
        \includegraphics[width=\linewidth]{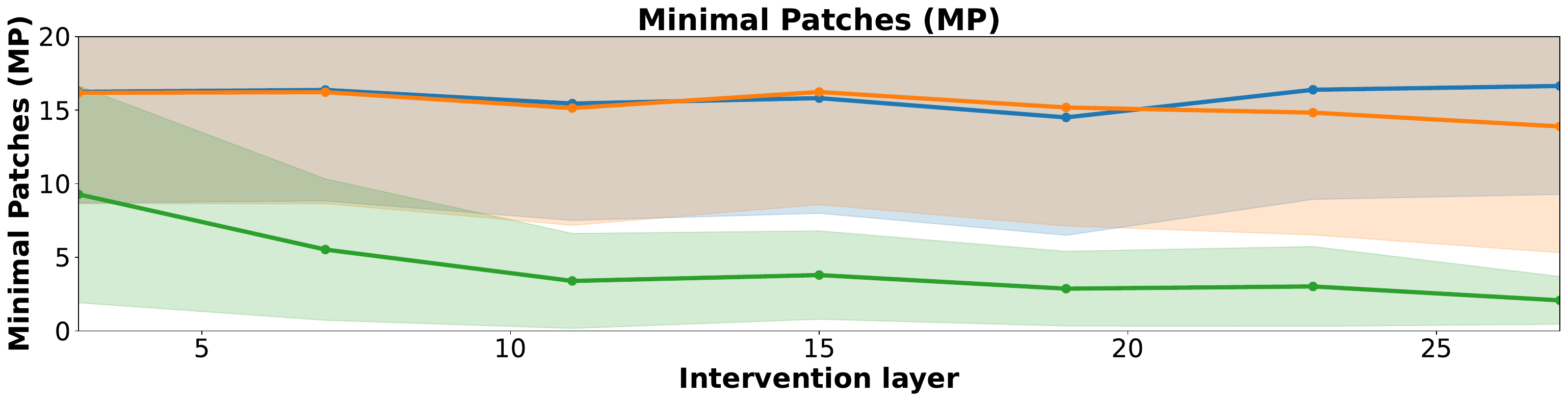}        
        
        \includegraphics[width=\linewidth]{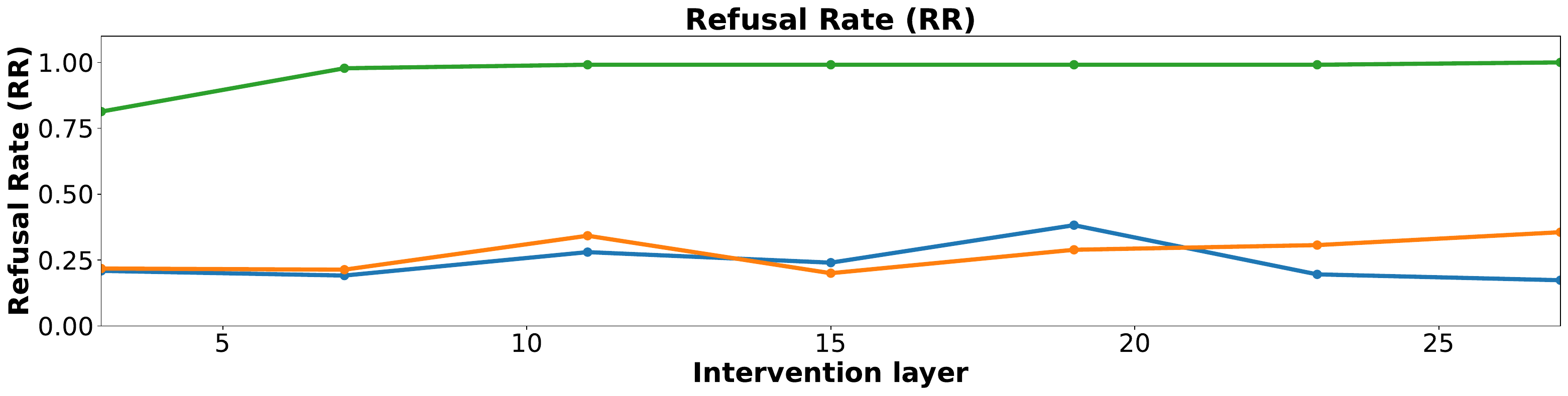}
        \caption{\textsc{Llama}}
        \label{fig:main-eval-llama}
    \end{subfigure}
    \hfill
    \begin{subfigure}[t]{0.45\textwidth}
        \centering
        \includegraphics[width=\linewidth]{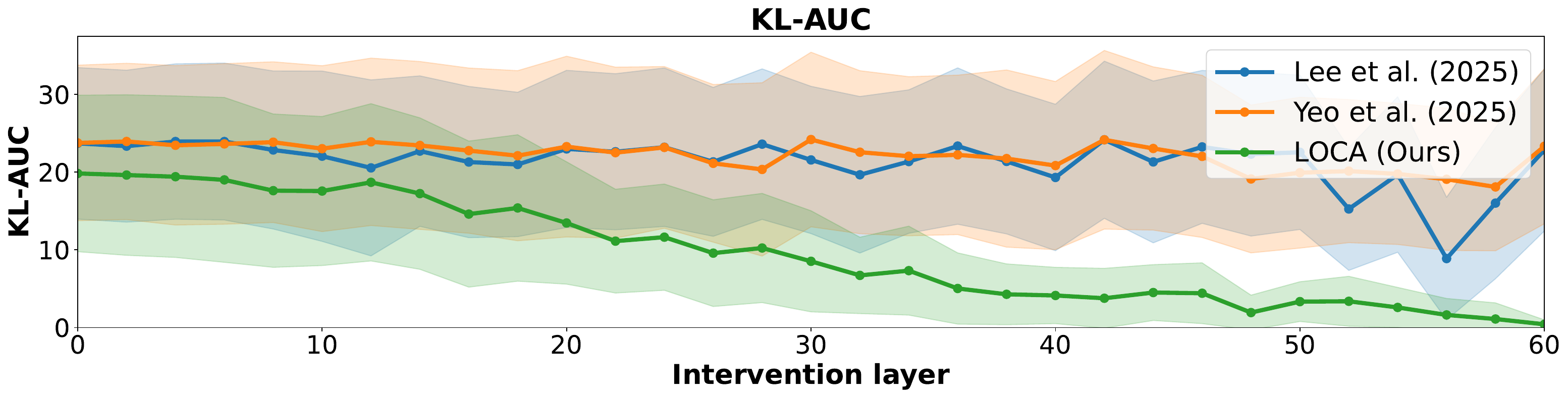}        
        
        \includegraphics[width=\linewidth]{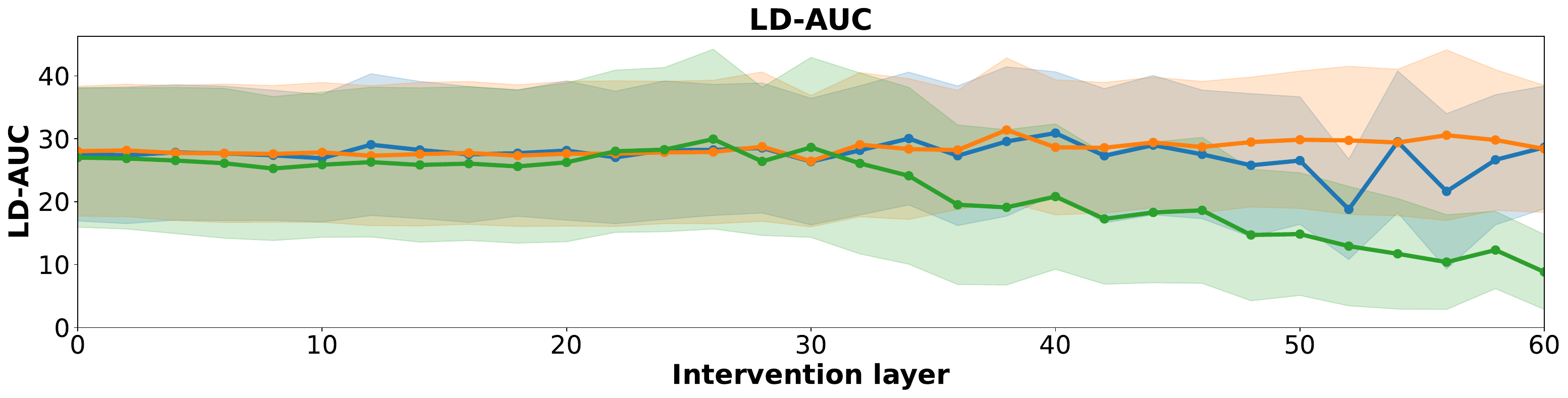}        
        
        \includegraphics[width=\linewidth]{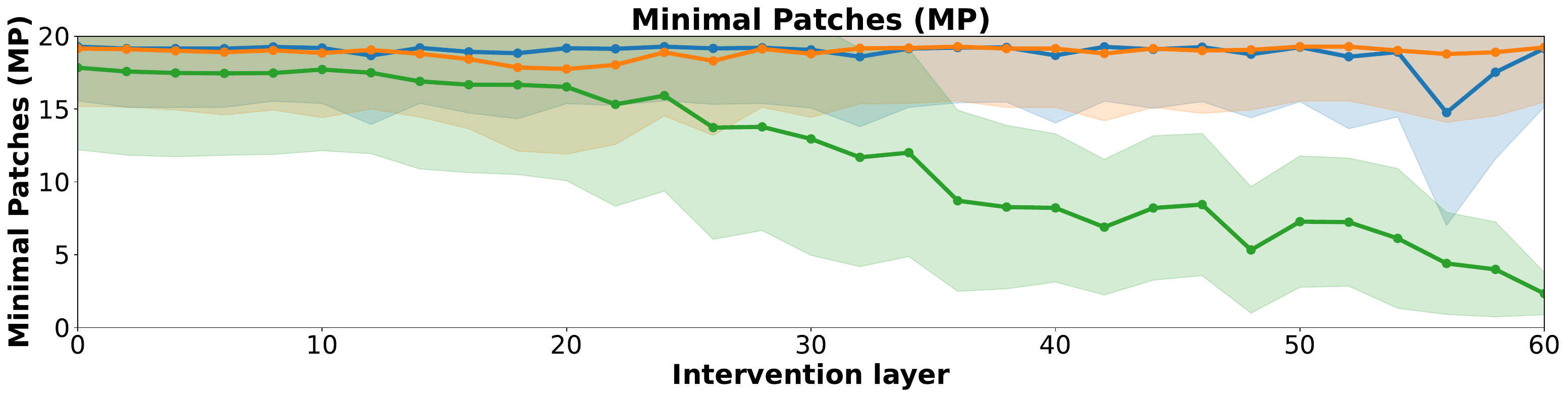}        
        
        \includegraphics[width=\linewidth]{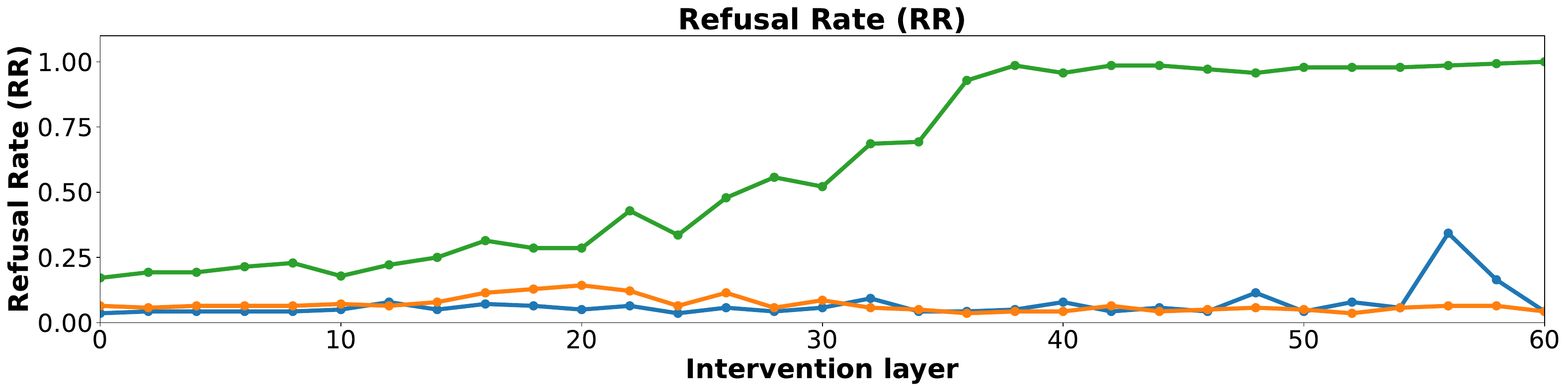}
        \caption{\textsc{Gemma-3}}
        \label{fig:main-eval-gemma}
    \end{subfigure}

    \caption{\textbf{LOCA induces refusal with minimal patches; other methods struggle.} We evaluate LOCA, \citet{lee2025upstreamrefusal}, and \citet{yeo-2025-understanding-refusal-saes} on 50 jailbreak requests and report KL-AUC, LD-AUC, MP, and RR on all layers (for which we have a pre-trained SAE available) on the \textsc{Gemma-3} and \textsc{Llama} models. All metrics are $\pm 1$ std. dev. except RR.}
    \label{fig:main-eval}
\end{figure}

\textbf{Metrics. } We are interested in measuring how closely the output from the patched embedding tokens matches the original output. To avoid generating the response after each successive patching operation, we propose metrics that measure differences between the original and patched output only for the \textit{first output token}. These metrics can be computed efficiently and serve as a proxy for how close the two generated outputs are. We introduce four metrics. After each patch operation, we compute \textbf{(1)} the KL Divergence (termed $KL$) of the two predicted next-token probabilities, and \textbf{(2)} the logit difference (termed $LD$) between the original and patched for the original's predicted token, as justified by \citet{Heimersheim2024ActPatchTutorial}. Since we are interested in achieving the lowest error with the smallest number of patches, we condense the metric to a scalar by reporting the Area Under the Curve (AUC) normalized by the total patches applied, which we term as $KL$-AUC and $LD$-AUC. To ensure we are able to recover refusal behavior with minimal changes, we record \textbf{(3)} the number of patches needed for the patched predicted token to match the first output token on the original prompt, termed \textit{Minimal Patches} (\textit{MP}). If refusal is not induced within some number of patches $K$, we set $MP= K$. After \textit{all} patches are applied, we calculate \textbf{(4)} the \textit{Refusal Rate} (\textit{RR}) by checking if the first output token after all patches matches the original first output token.\footnote{Ideally, refusal would be measured on the entire generated response, not the first token. To supplement our first-token proxy metrics, Appendix \ref{sec:appendix-evaluating-past-first-token-proxy} evaluates a subset of \textsc{Llama} samples using the Harmbench autograder. We note failure cases of the first-token proxy in Appendix \ref{sec:appendix-failure-cases}.}

\begin{figure}[t]
    \centering

    \begin{subfigure}[t]{0.48\textwidth}
        \centering
        \includegraphics[width=\linewidth]{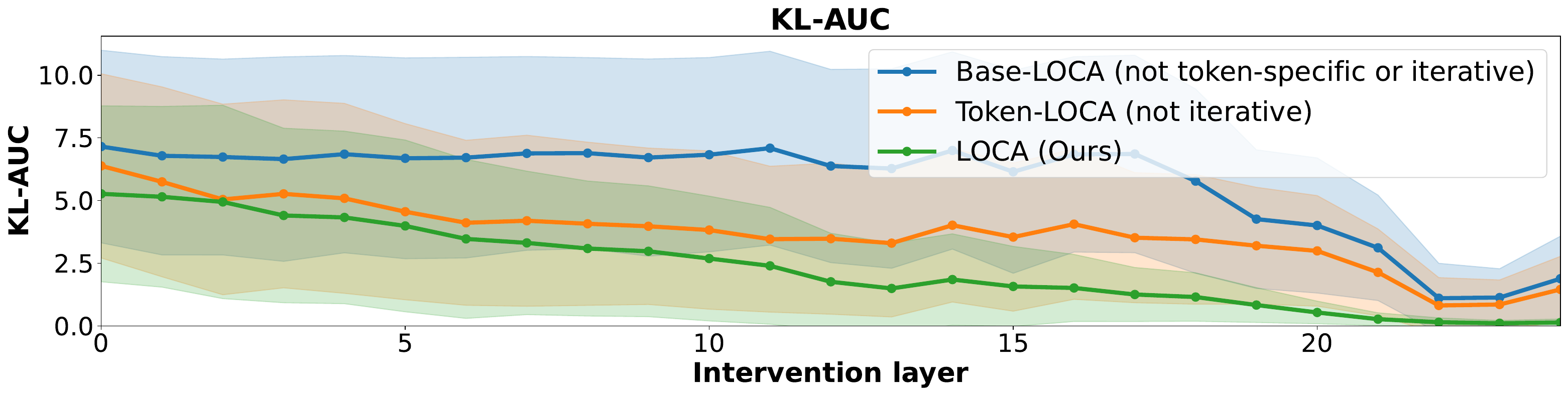}        
        
        \includegraphics[width=\linewidth]{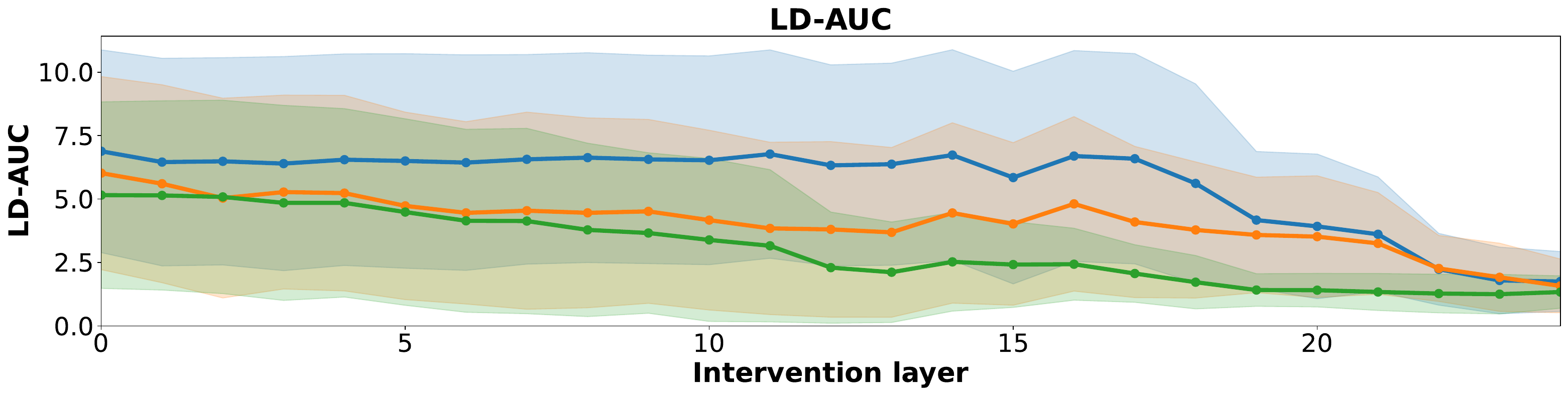}        
        
    \end{subfigure}
    \hfill
    \begin{subfigure}[t]{0.48\textwidth}
        \centering
        \includegraphics[width=\linewidth]{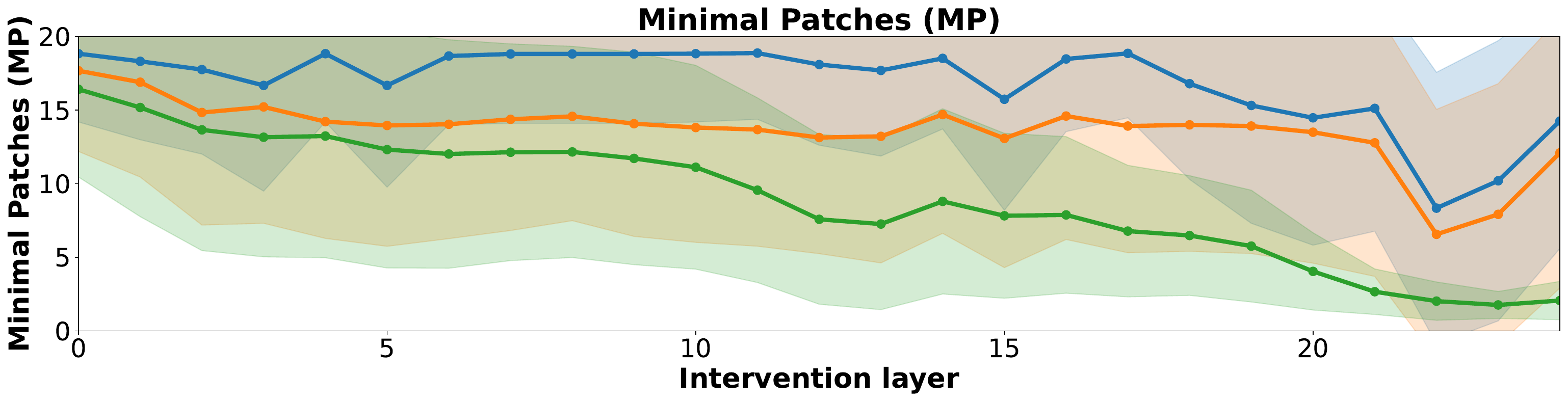}        
        
        \includegraphics[width=\linewidth]{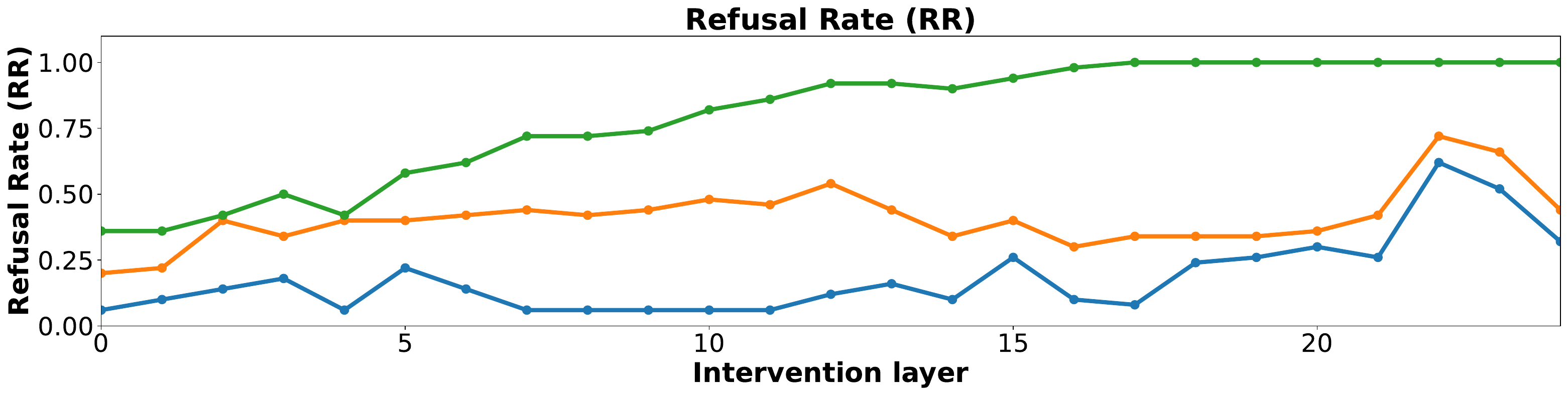}        
        
    \end{subfigure}

    \caption{\textbf{LOCA is better because it is token-specific and iterative.} We ablate key aspects of LOCA by creating two variants. Base-LOCA is neither token-specific nor iterative. Token-LOCA is token-specific, but not iterative. We evaluate on \textsc{Gemma-2}, with the same setup as Sec. \ref{sec:main-experiment}. Prior works are variants of Base-LOCA, which explains their poor performance.}
    \label{fig:ablation-eval}
\end{figure}

\subsection{LOCA generates minimal, local, causal explanations of jailbreak success}
\label{sec:main-experiment}
We perform model-specific filtering of the test set, keeping original-jailbreak pairs where $x_o$ failed and $x_j$ succeeded (as measured by Harmbench).\footnote{Since not all jailbreak methods are not equally successful at jailbreaking the LLM, the evaluation set does not represent a balanced coverage of all jailbreak types in the WhatFeatures dataset. We include the evaluated jailbreak method distribution in Appendix \ref{sec:appendix-distribution-of-jailbreak-prompts}.} This yields $225$ and $438$ eligible evaluation samples for \textsc{Llama} and \textsc{Gemma-3}, respectively.\footnote{Given computational constraints, we evaluate a set of 140 samples for \textsc{Gemma-3}.} We test up to $K=20$ interventions from each method across all intermediate layers (except the last layer) for which a corresponding SAE exists. We report the \textit{KL-AUC}, \textit{LD-AUC}, \textit{MP}, and \textit{RR} for \textsc{Gemma-3} and \textsc{Llama} in Fig. \ref{fig:main-eval}. Results for \textsc{Qwen} and \textsc{Gemma-2} are in Appendix \ref{sec:appendix-more-main-results}.

\textbf{Astonishingly, we find that LOCA can, on average, induce refusal behavior on \textsc{Llama} with 6-8 patching operations on early-layer residual embeddings}. This is a significant reduction from prior work in activation steering \citep{arditi2024refusal}, which induced refusal by steering mid-layer residual embeddings along \textit{every} input token position (which, depending on the input, could be from 50-200 tokens). For \textsc{Gemma-3}, RR increases with layer depth, with success being reached in intermediate layers (layer 36 onwards) with an average of 9 patching operations. Less patching operations are needed in downstream layers.

LOCA substantially outperforms the baselines across our other metrics. As the layer on which LOCA performs the patching increases, our per-patching metrics (\textit{LD-AUC} and \textit{KL-AUC}) decrease. \textit{MP} conclusively proves that LOCA is able to recover the original prompt output in fewer patching operations, and LOCA's \textit{RR} consistently increases for both \textsc{Llama} and \textsc{Gemma-2}, reaching 100\% at layers 7 and 38, respectively. Prior work struggles to improve \textit{RR} as the intervention layer increases, and they likewise fail to achieve improvements on the other metrics. We explore reasons for this next. 

\subsection{LOCA succeeds by making iterative, token-specific changes}
LOCA makes two distinctive improvements over prior work. \textbf{First, LOCA computes the first-order approximation for \textit{each token}}, while prior works average the gradient over the entire token span. Second, \textbf{LOCA uses an iterative algorithm}, where the approximation is recalculated after applying each patch operation to take token interaction effects into account.\footnote{While this means LOCA's compute time scales linearly with the number of iterations, the improvements are empirically justified. A comparison on computation time is in Appendix \ref{sec:appendix-compute}.} To validate the effect of these improvements, we create two variants of LOCA and repeat the experiment from Section \ref{sec:main-experiment}. Base-LOCA computes the \textit{gradient term} for the first-order approximation by averaging over the tokens. The magnitude term is still token-specific. Token-LOCA retains the token-specific first-order approximation of LOCA, but does not use an iterative algorithm. The ordering is calculated once and used to identify the top-$K$ patching operations. Results are reported in Fig \ref{fig:ablation-eval} for \textsc{Gemma-2}. Across all metrics, LOCA outperforms the variants. Prior methods from \citet{lee2025upstreamrefusal, yeo-2025-understanding-refusal-saes} can be seen as variants of Base-LOCA with different objective functions; thus, these results help explain their shortcomings in generating minimal, local, causal explanations.

\begin{figure*}[t]
    \centering
    \setlength{\tabcolsep}{3pt}

    \begin{tabular}{
        >{\centering\arraybackslash}m{0.05\textwidth}
        >{\centering\arraybackslash}m{0.3\textwidth}
        >{\centering\arraybackslash}m{0.3\textwidth}
        >{\centering\arraybackslash}m{0.3\textwidth}
    }
        &
        \textbf{\small Early} &
        \textbf{\small Early-Middle} &
        \textbf{\small Middle} \\

        \rotatebox{90}{\small Location} &
        \includegraphics[width=\linewidth]{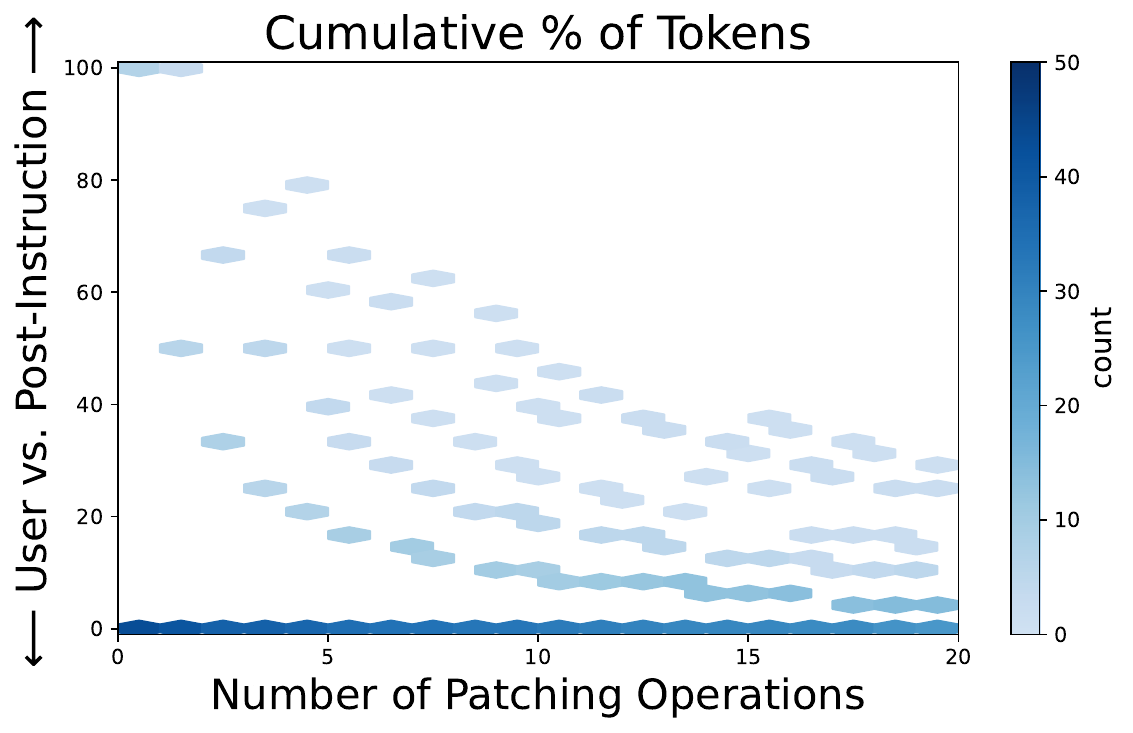} &
        \includegraphics[width=\linewidth]{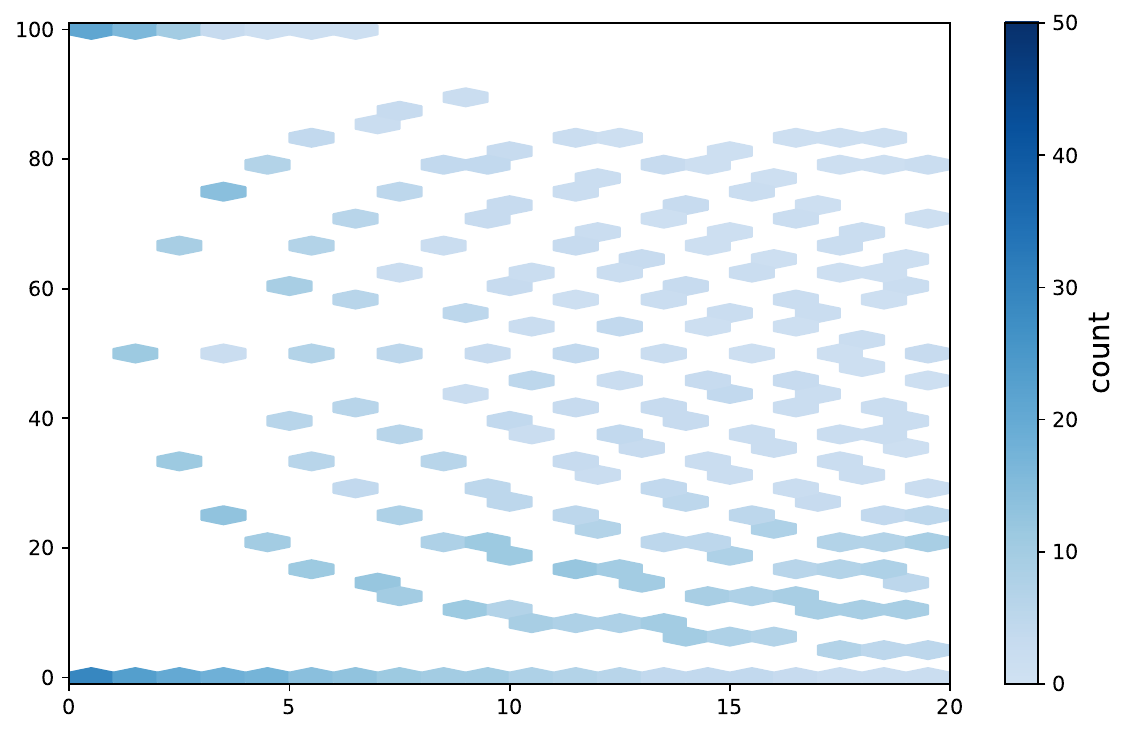} &
        \includegraphics[width=\linewidth]{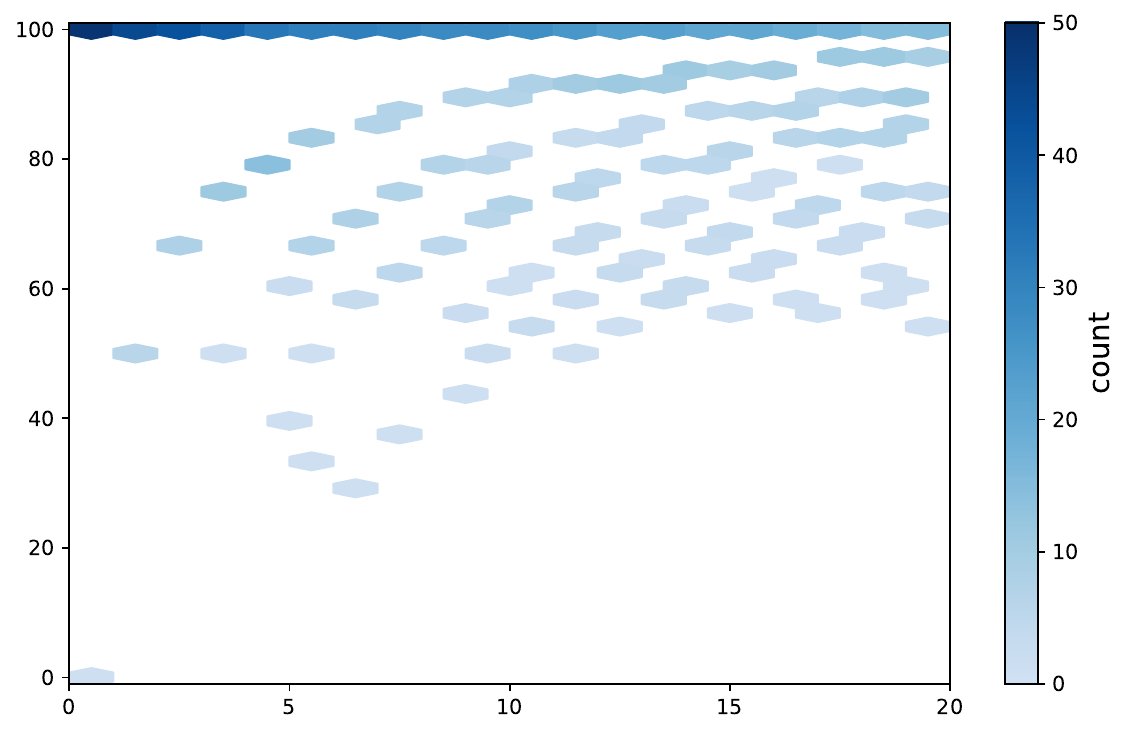} \\

        \rotatebox{90}{\small Token} &
        \includegraphics[width=\linewidth]{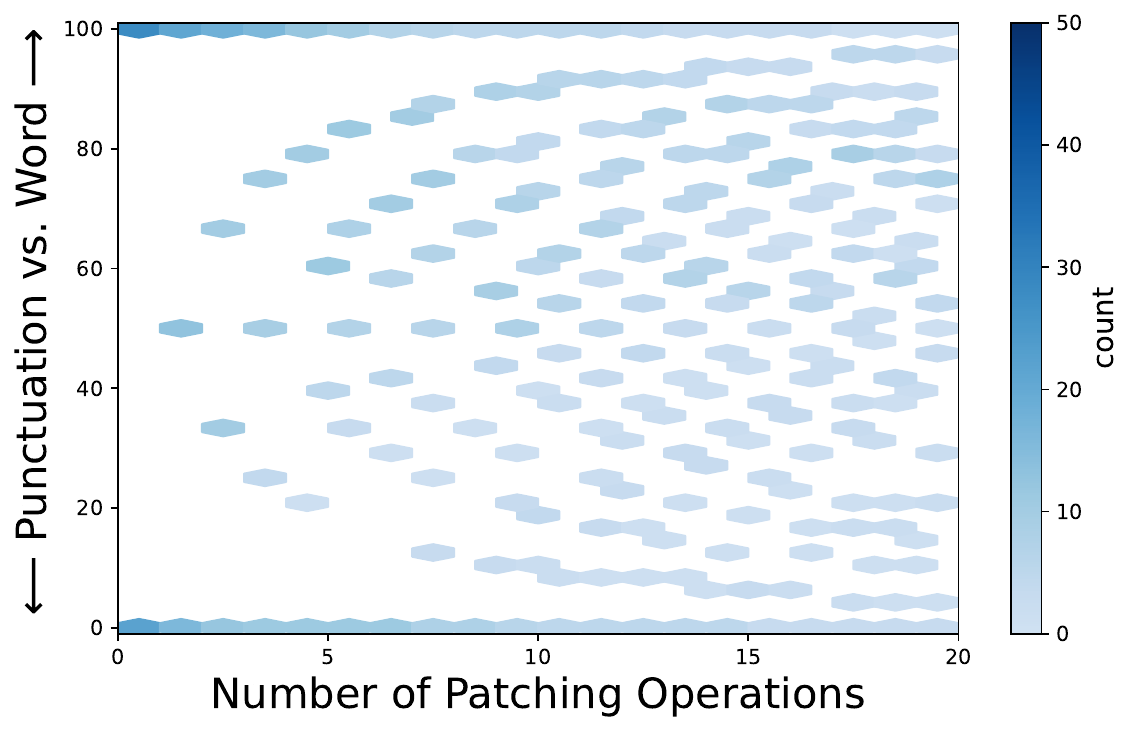} &
        \includegraphics[width=\linewidth]{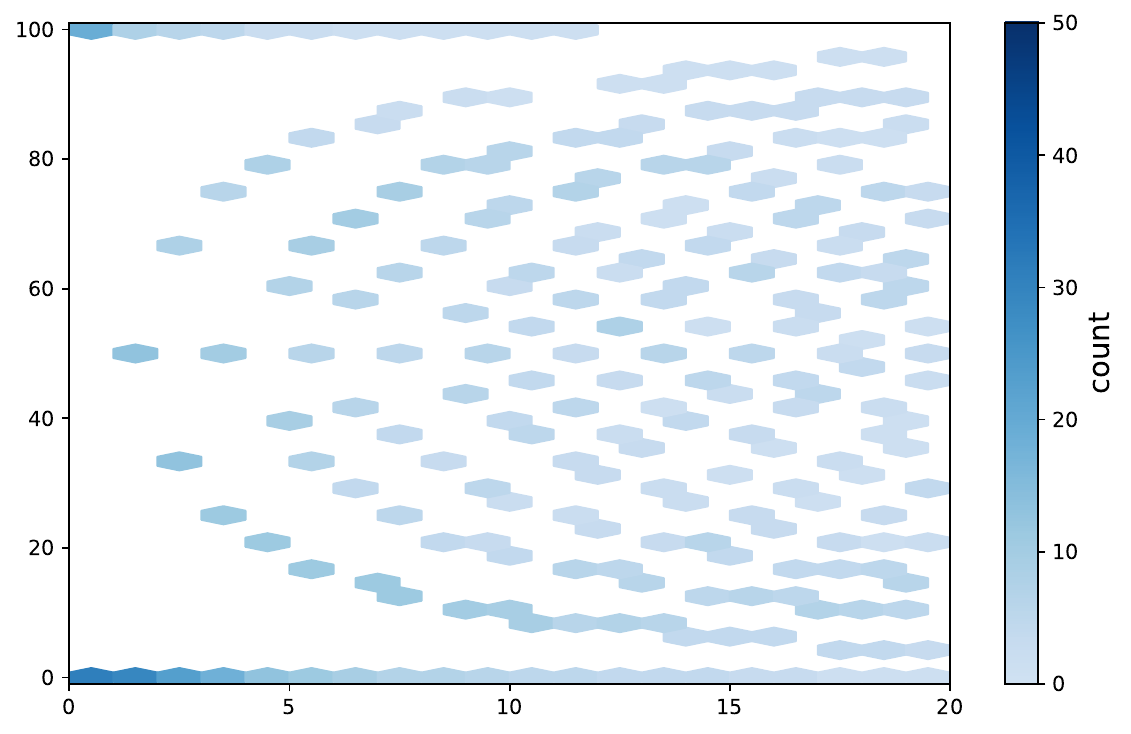} &
        \includegraphics[width=\linewidth]{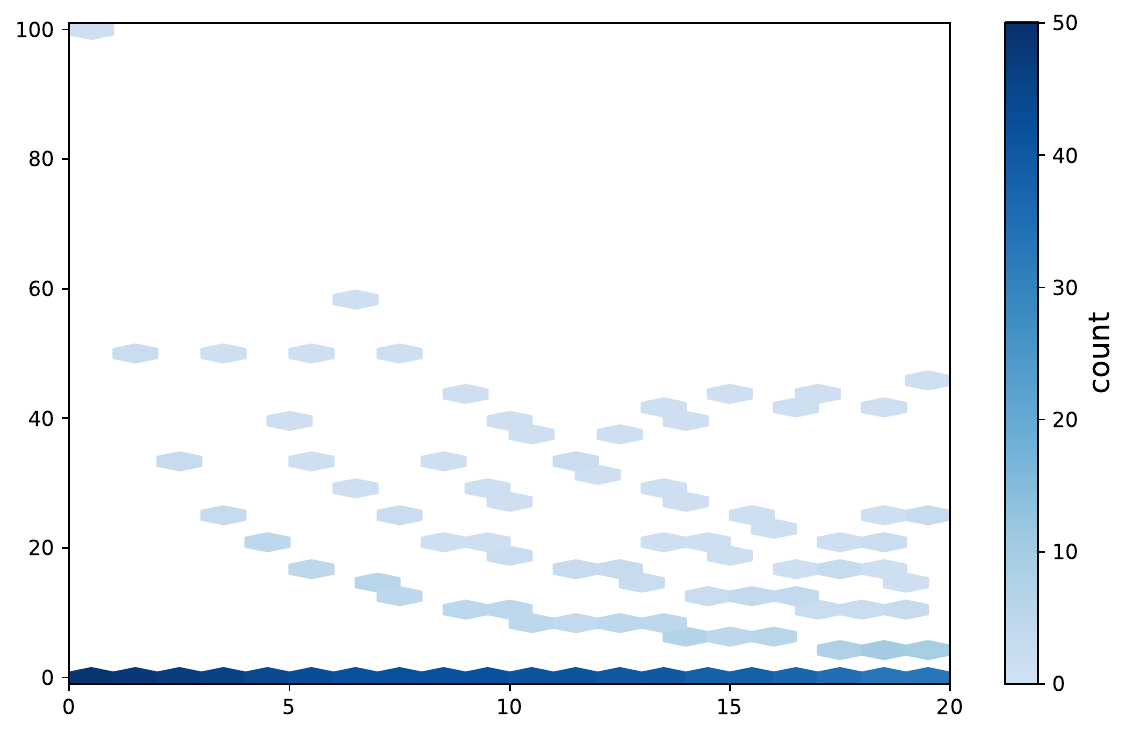}
    \end{tabular}

    \caption{\textbf{Localization analysis.} We analyze the tokens (both location and type) LOCA selects for \textsc{Llama} at three layer depths. (Top) LOCA tends to select user instruction tokens in early layers, and post-instruction tokens in later layers. (Bottom) Early layers do not have a bias towards any token type; later layers almost exclusively chose \textsc{punctuation} tokens.}
    \label{fig:location-type-analysis}
\end{figure*}

\subsection{Where and which tokens are most important for explaining jailbreak success?}
\label{sec:loca-localization-experiment}

Existing work shows that ending tokens (final instruction token and post-instruction tokens) causally represent harmfulness and refusal \citep{bau_refusal_and_harm_seperate}. However, the role of the other instruction tokens is unclear. Given that the refusal signal is represented as a low-dimensional subspace in the residual stream of the middle layers, one hypothesis is that earlier layers perform computations over the input prompt to determine the middle-layer refusal signal \citep{lee2025upstreamrefusal}. Given that LOCA can find minimal, causal interventions to induce refusal, we add additional insight to this hypothesis by studying which tokens (in terms of both location and type) LOCA selects for patching. We show results for \textsc{Llama} here; results for \textsc{Gemma-2}, which are similar, are in Appendix \ref{sec:appendix-localization}.

To study token location, we create a distribution plot (top row of Fig \ref{fig:location-type-analysis}) of the cumulative percentage of tokens LOCA selected that belong to \textcolor{blue}{$T_{\text{post-inst}}$} (as opposed to \textcolor{orange}{$T_{\text{inst}}$}) at three layer depths: (i) early, (ii) early-middle, (iii) middle.\footnote{For \textsc{Llama}, we analyze 3, 7, and 15. For \textsc{Gemma-2}, we analyze layers 5, 10, and 15.} In early layers, LOCA tends to select \textcolor{orange}{$T_{\text{inst}}$} tokens. In early-middle layers, the location of selected tokens does not have a strong bias, but in middle layers, the selection skews towards \textcolor{blue}{$T_{\text{post-inst}}$} tokens, while keeping a non-negligible distribution of \textcolor{orange}{$T_{\text{inst}}$} tokens.

We study token type by classifying tokens as either \textsc{punctuation} or \textsc{word}. A token is a \textsc{word} if it starts with a letter or contains a number or decimal character. Otherwise, the token is a \textsc{punctuation}. Given their syntactic role in the chat template, all tokens in \textcolor{blue}{$T_{\text{post-inst}}$} are classified as \textsc{punctuation}. The \textsc{word} vs. \textsc{punctuation} distribution plot is shown in the bottom row of Fig \ref{fig:location-type-analysis}. For both models, early and early-middle layers do not exhibit a strong bias towards either token type. Middle layers exhibit a strong shift to \textsc{punctuation} tokens, in accordance with the shifted emphasis to \textcolor{blue}{$T_{\text{post-inst}}$} tokens. 

We add additional insight to the earlier hypothesis: the refusal signal is determined mostly by \textcolor{orange}{$T_{\text{inst}}$} tokens in early layers, with no preference for token type. In middle layers, refusal is influenced more by \textcolor{blue}{$T_{\text{post-inst}}$} tokens, but \textit{can still be present} in \textsc{punctuation} \textcolor{orange}{$T_{\text{inst}}$} tokens.

\subsection{Case study: how does \texttt{AutoDAN} convince \textsc{Llama} to give instructions on illegally acquiring firearms?}
\label{sec:case-study}
Given that LOCA can find explanations on \textsc{Llama} that require only a few activation patches, we examine one explanation as a case study.\footnote{We anecdotally find interpreting \textsc{Llama} significantly easier than \textsc{Gemma-2}. We attribute this to the availability of a \textsc{Llama} pre-trained SAE on chat data, which surfaces more conversational and jailbreak-relevant concepts than GemmaScope, which was trained on general pre-training data.} An \texttt{AutoDAN} jailbreak \citep{liu2024autodan}, where the model is told to operate in a privileged ``Developer Mode", is effective at eliciting harmful responses from \textsc{Gemma-2} and \textsc{Llama} models. To understand why, we apply LOCA to a successful \texttt{AutoDan}-jailbroken prompt asking for instructions on acquiring illegal firearms. Full outputs from the LOCA algorithm are in Appendix \ref{sec:appendix-case-study}; SAE concepts are interpreted using Neuronpedia. After two activation patches in a middle layer (layer 11), LOCA induces refusal behavior. The first patch is on the \texttt{assistant} \textcolor{blue}{$T_{\text{post-inst}}$} token; the patch increases the strength of concept \#31126, which activates on harmful requests (e.g., violence, sexual content, illegal behavior). The second activation patch is applied on a \textsc{punctuation} \textcolor{orange}{$T_{\text{inst}}$} token. The patch decreases concept \#125009, which maximally activates on harmless requests concerning code generation. Thus, we interpret \texttt{AutoDAN}'s jailbreak success through two mechanisms: (1) a refusal concept associated with harmful prompts was suppressed and (2) the jailbreak convinced the model that operating in ``Developer Mode" was akin to answering harmless questions on code generation. 

Next, we apply LOCA on an early layer (layer 3) for a more fine-grained understanding. LOCA induces refusal after five activation patches, with most of them concentrating on \textsc{punctuation} tokens. The first activation patch is on a ``fabricating" \textcolor{orange}{$T_{\text{inst}}$} token, which instructs the model to compensate for a lack of knowledge by fabricating answers. Interestingly, the patch decreases concept \#21337, which activates on generic text. To answer this specific prompt, the model must provide website URLs to acquire illegal firearms, and we interpret the first patching operation as evidence that the model believes fabricating this information is harmless. Follow-up patches mainly decrease the strength of harmless, continuation-related concepts. This supports the view that the jailbreak succeeds by increasing the strength of harmless concepts in key \textsc{punctuation} and \textcolor{blue}{$T_{\text{post-inst}}$} tokens. Unintuitively, the final activation patch increases a concept that maximally activates on generally harmless questions, which leads to LOCA inducing refusal. It is likely we are unable to interpret the final concept correctly.


\section{Limitations \& Conclusion}

LOCA is a step towards creating minimal, local, causal explanations of jailbreak success. However, it is not without issues; in particular, Appendix \ref{sec:appendix-failure-cases} details failure cases we encountered during manual inspection. Future work should seek to improve on the first-token proxy for refusal. Furthermore, interpreting LOCA explanations on earlier layers of \textsc{Llama} was time-intensive and error prone, which we largely attribute to limitations in SAEs. While SAEs can provide interpretable directions, it's possible that these directions are not ``optimal" for jailbreak explanations. We are excited by recent developments in Neural Geometry \citep{geigerNeuralGeometryGoodfire, fel2026structuring}, where modeling jailbreak relevant concepts as non-linear manifolds may ease the interpretation difficulties. Finally, the ideas from this work---specifically, the shift to local explanations and token-specific patching---can be used to form the foundation of robust inference-time, steering-based defenses to jailbreaks.

\section*{Author Contributions}
Shubham Kumar conceived the research problem, developed the proposed method, designed and conducted the experiments, analyzed and visualized the results, and wrote the manuscript. Narendra Ahuja supervised the research and provided guidance and feedback throughout the project and during preparation of the manuscript.

\section*{Acknowledgments}
The support of the IBM-Illinois Discovery Accelerator Institute (IIDAI), Office of Naval Research under grant N00014-24-1-2169, Amazon–Illinois Center on AI for Interactive Conversational Experiences, and USDA National Institute of Food and Agriculture under grant AFRI 2020-67021-32799/1024178 are gratefully acknowledged. This material is based upon work supported by the National Science Foundation Graduate
Research Fellowship Program under Grant No. DGE 21-46756. Any opinions, findings,
and conclusions or recommendations expressed in this material are those of the author(s)
and do not necessarily reflect the views of the National Science Foundation. Shubham Kumar also gratefully acknowledges the support of UIUC ECE's Distinguished Research Fellowship and Promise of Excellence Fellowship. This work used Delta at NCSA through allocation CIS250999 from the Advanced Cyberinfrastructure Coordination Ecosystem: Services \& Support (ACCESS) program, which is supported by U.S. National Science Foundation grants \#2138259, \#2138286, \#2138307, \#2137603, and \#2138296.

\bibliography{colm2026_conference}
\bibliographystyle{colm2026_conference}

\appendix
\clearpage

\section{Distribution of Evaluated Jailbreak Methods}
\label{sec:appendix-distribution-of-jailbreak-prompts}

\begin{figure}[h]
    \centering

    \begin{subfigure}[t]{0.49\textwidth}
        \centering
        \includegraphics[width=\linewidth, valign=t]{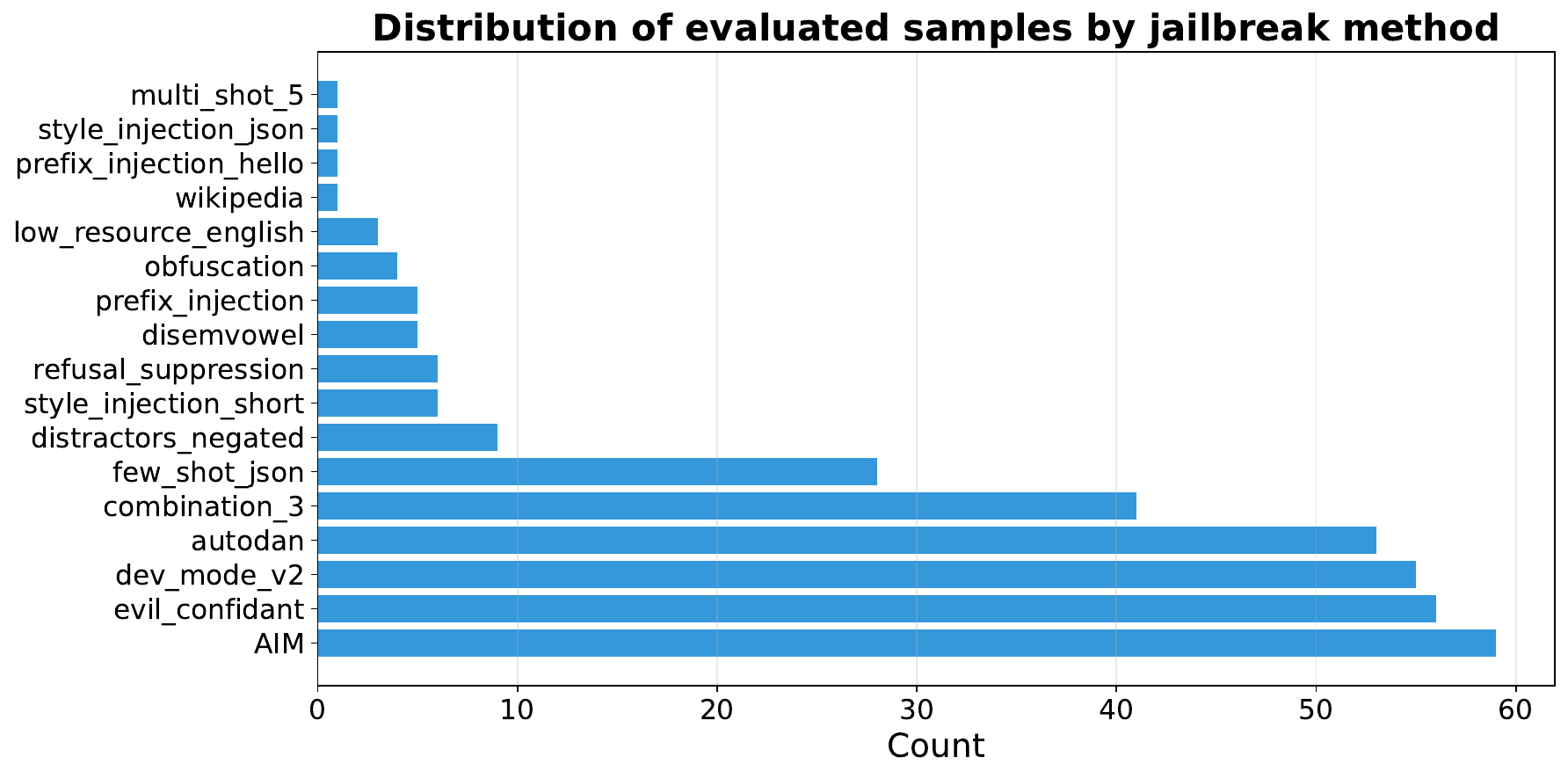}
        \caption{\textsc{Gemma-2}}        
    \end{subfigure}
    \hfill
        \begin{subfigure}[t]{0.49\textwidth}
        \centering
        \includegraphics[width=\linewidth, valign=t]{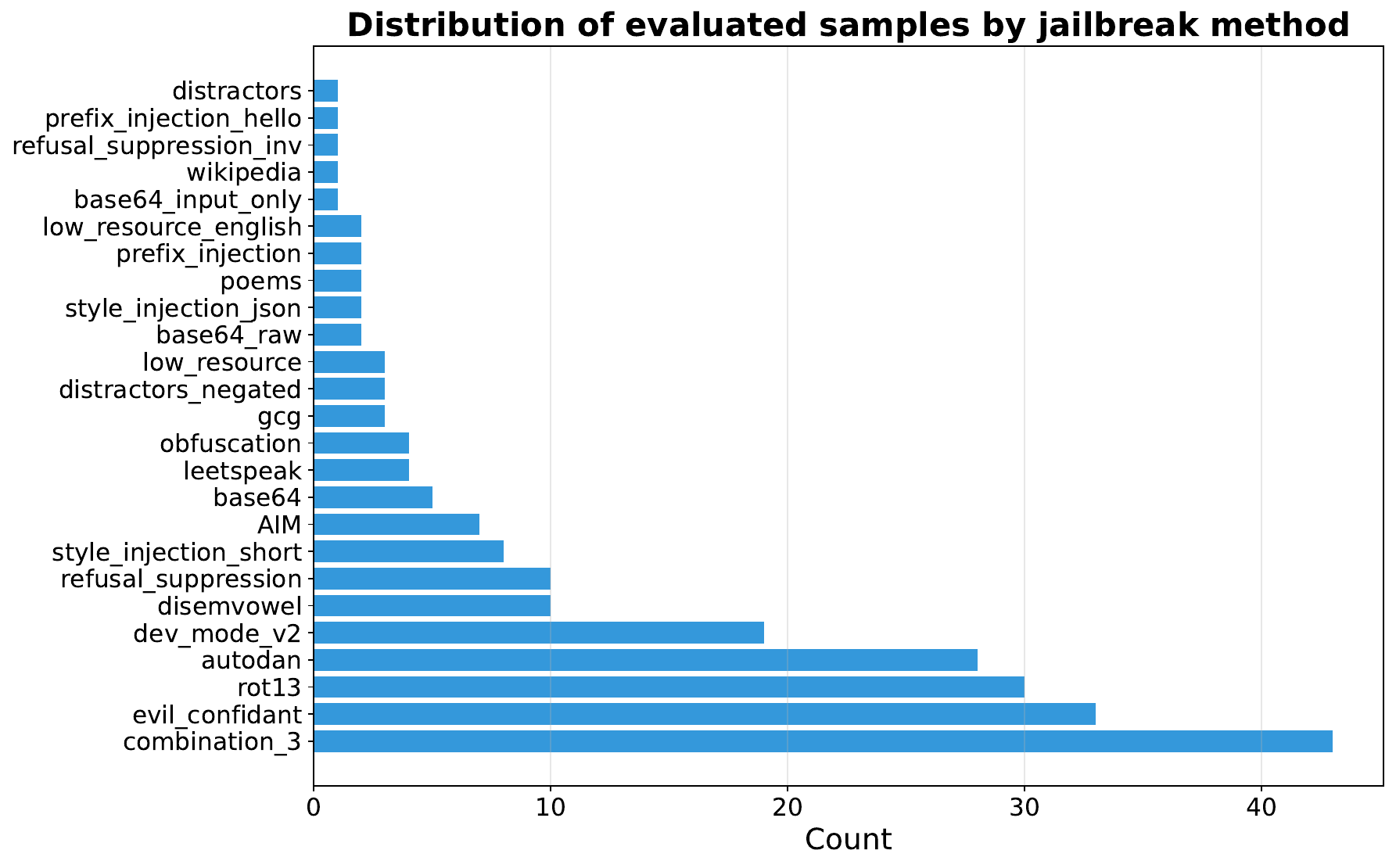}
        \caption{\textsc{Llama}}        
    \end{subfigure}

    \caption{Distribution of the jailbreak methods used for evaluation in our experiments for \textsc{Llama} and \textsc{Gemma-2}. Observe that the distribution across jailbreak types is not uniform.}
    \label{fig:appendix-jailbreak-distribution}
\end{figure}

We only apply LOCA to samples where the original, harmful prompt failed but the corresponding jailbreak prompt succeeded. This filtering is model-specific and does not result in a balanced coverage of the prompt types in the WhatFeatures set, since not all prompt types are equally successful at jailbreaking the model. We include the prompt method distribution for \textsc{Llama} and \textsc{Gemma-2} in Fig. \ref{fig:appendix-jailbreak-distribution}.

\FloatBarrier

\section{Justifying The First-Token Proxy for Refusal Behavior}
\label{sec:appendix-justifying-first-token-proxy}
This paper opts to reduce refusal to a single output token because this results in an efficient proxy for optimization and evaluation.

\begin{figure}[t]
    \centering

    \begin{subfigure}[t]{0.49\textwidth}
        \centering
        \includegraphics[width=\linewidth, valign=t]{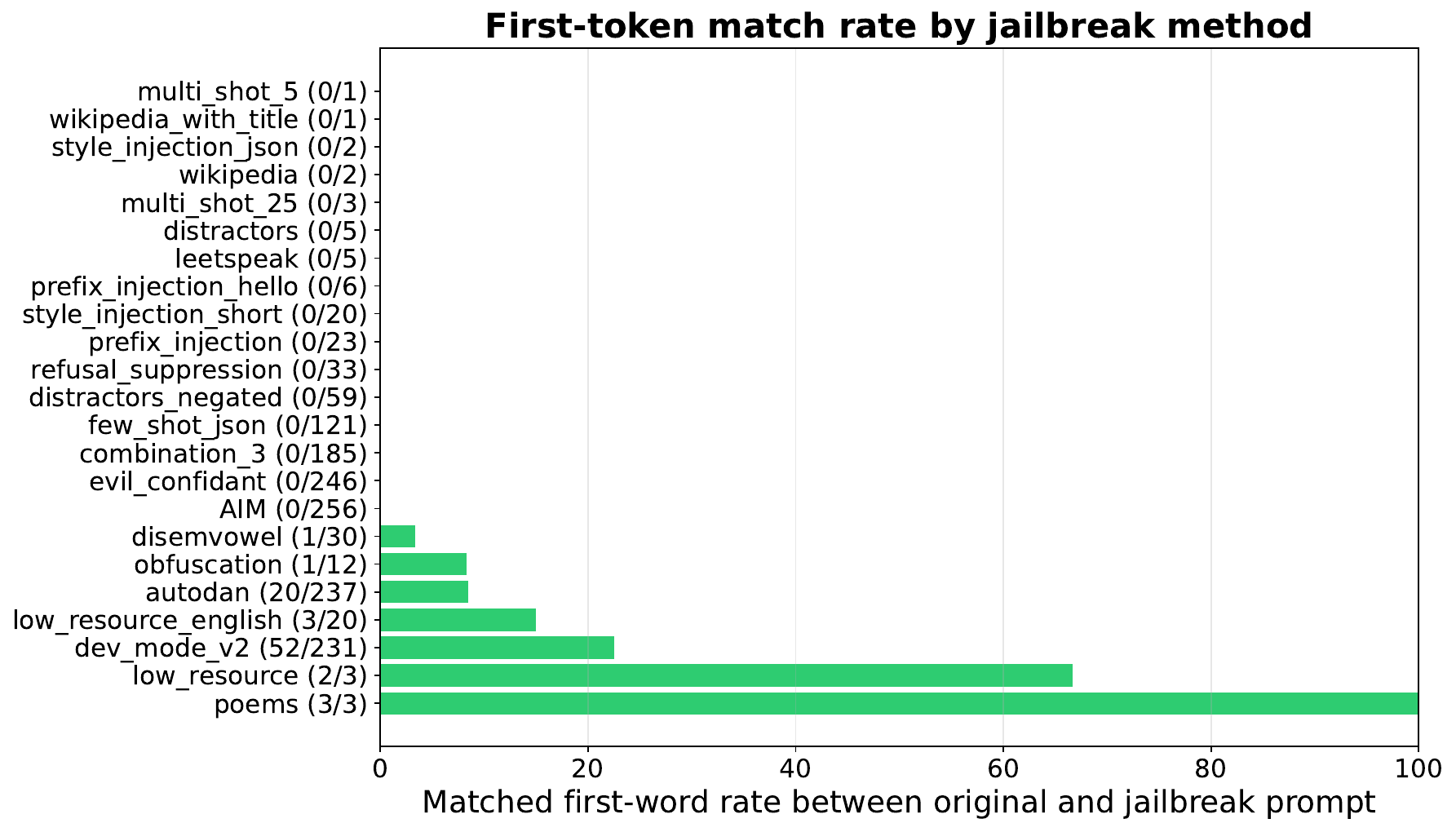}
        \caption{\textsc{Gemma-2}}        
    \end{subfigure}
    \hfill
        \begin{subfigure}[t]{0.49\textwidth}
        \centering
        \includegraphics[width=\linewidth, valign=t]{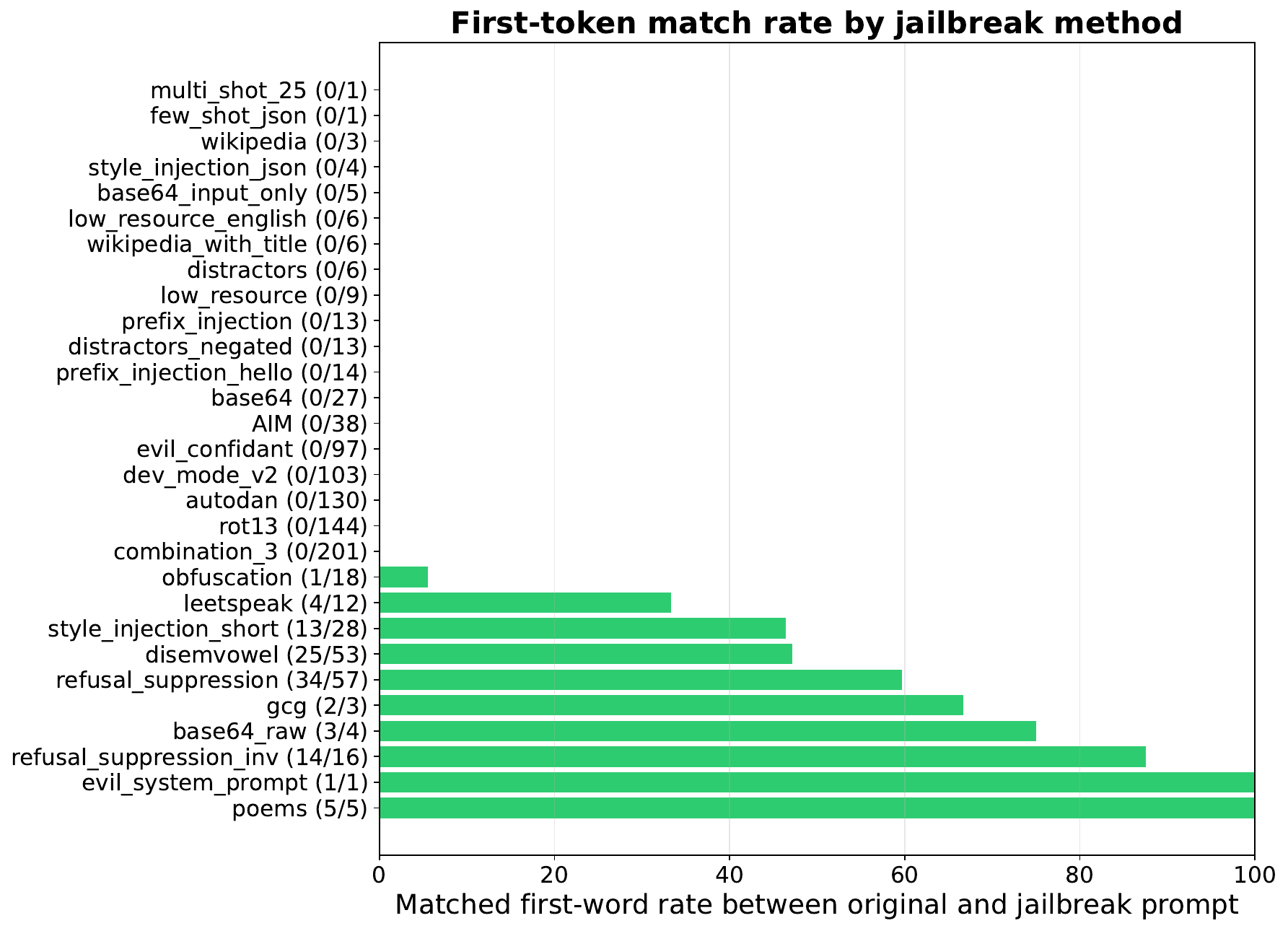}
        \caption{\textsc{Llama}}        
    \end{subfigure}

    \caption{For each jailbreak type, we show the frequency of jailbreak prompts that have the same first output token as the corresponding original, harmful prompt. Observe that most jailbreaks types do not have the same first output token. For those that do, it's typically highly prevalent.}
    \label{fig:appendix-analysis-of-first-token}
\end{figure}

To justify this, we find the distribution of eligible samples across the WhatFeatures (train and test) where the LLM’s first output word is the same (normalized for capitalization, punctuation, and whitespace) for corresponding original and jailbreak prompts. For \textsc{Llama}, there are 102/1018 such samples (10.0\%), and for \textsc{Gemma-2}, there are 82/1504 such samples (5.5\%). To understand the per-jailbreak-method breakdown, we plot the frequency of how often this occurs in Fig. \ref{fig:appendix-analysis-of-first-token}. We find that for most jailbreak methods, this occurs seldomly; for a few methods, it occurs very often. The first-token proxy is certainly a limitation of our approach, and we show failure modes resulting from the proxy in Appendix \ref{sec:appendix-failure-cases}. However, we highlight that the limitation is \textit{anticipatable}---\textbf{we largely know for which prompt types the failure is likely to occur before running LOCA}. Considering this, along with the overall low frequency of violating samples (10\%), we believe that our use of a first-token proxy as an objective is justified. 

Future work may study how to alter the first-token optimization objective to mitigate this limitation. One idea is to minimize the LOCA objective over the first $N$ output token positions.

\section{Main Experiment Results on Additional Models}
\label{sec:appendix-more-main-results}

Due to space constraints, additional results for \textsc{Gemma-2} and \textsc{Qwen} are presented here. We follow all procedures outlined for the main experiment in Sec. \ref{sec:main-experiment}.

\begin{figure}[t]
    \centering

    \begin{subfigure}[t]{0.49\textwidth}
        \centering
        \includegraphics[width=\linewidth]{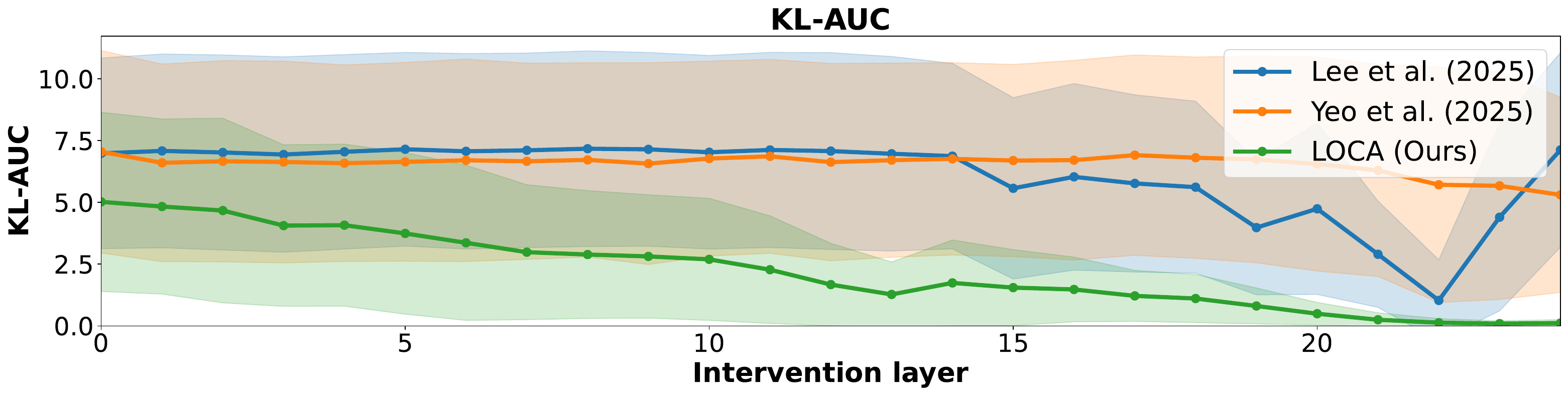}        
        
        \includegraphics[width=\linewidth]{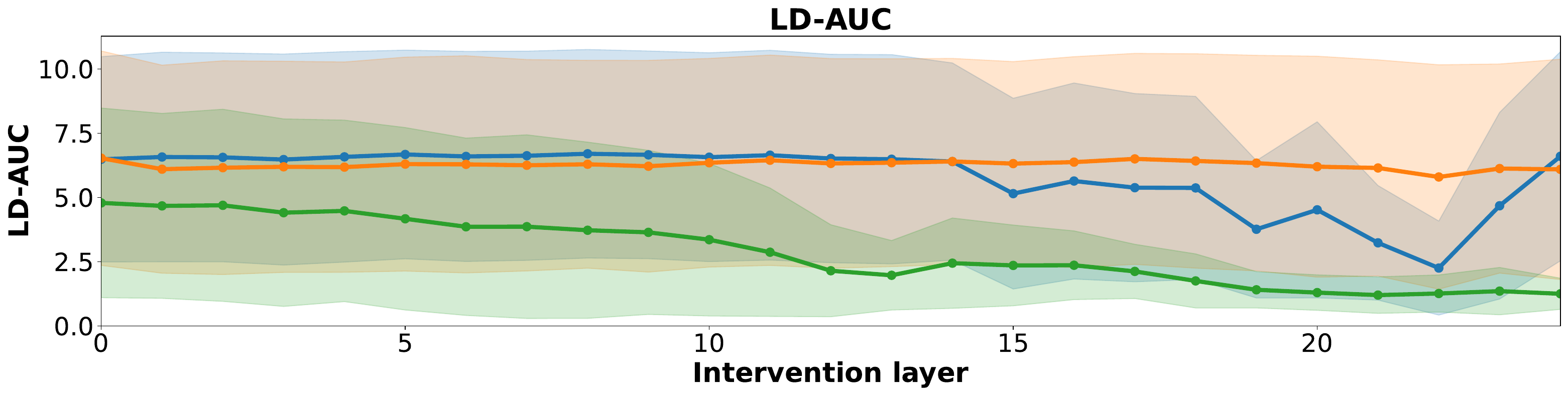}        
        
        \includegraphics[width=\linewidth]{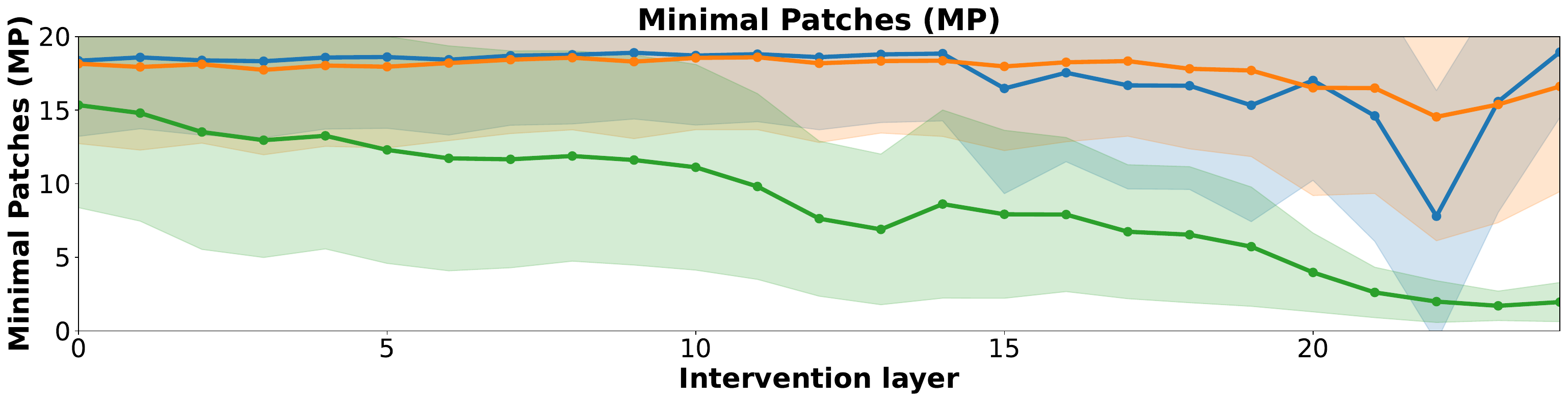}        
        
        \includegraphics[width=\linewidth]{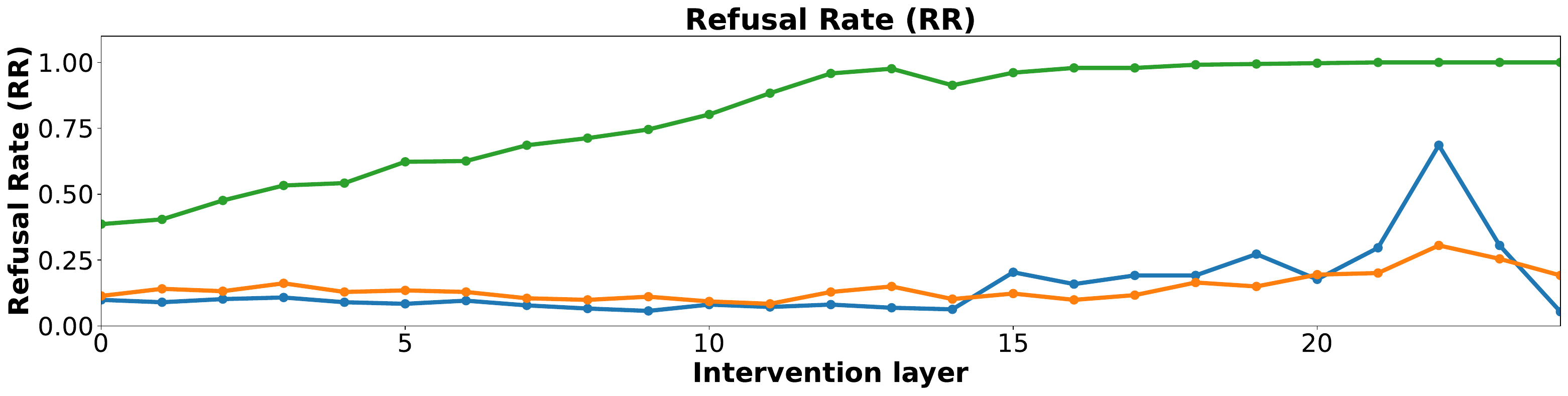}
        \caption{\textsc{Gemma-2}}
        \label{fig:main-eval-gemma-2}
    \end{subfigure}
    \hfill
        \begin{subfigure}[t]{0.49\textwidth}
        \centering
        \includegraphics[width=\linewidth]{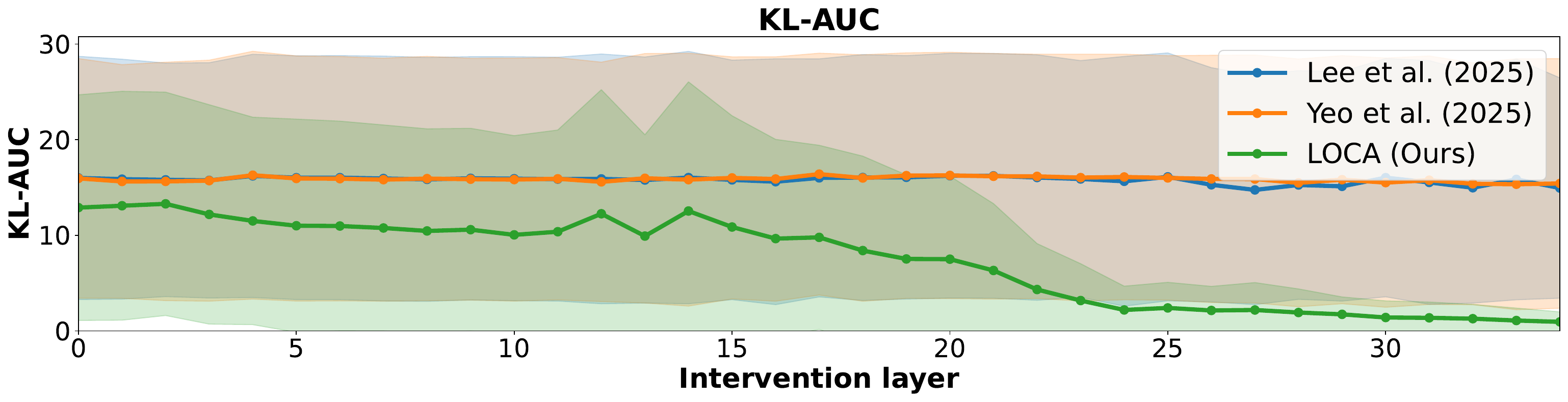}        
        
        \includegraphics[width=\linewidth]{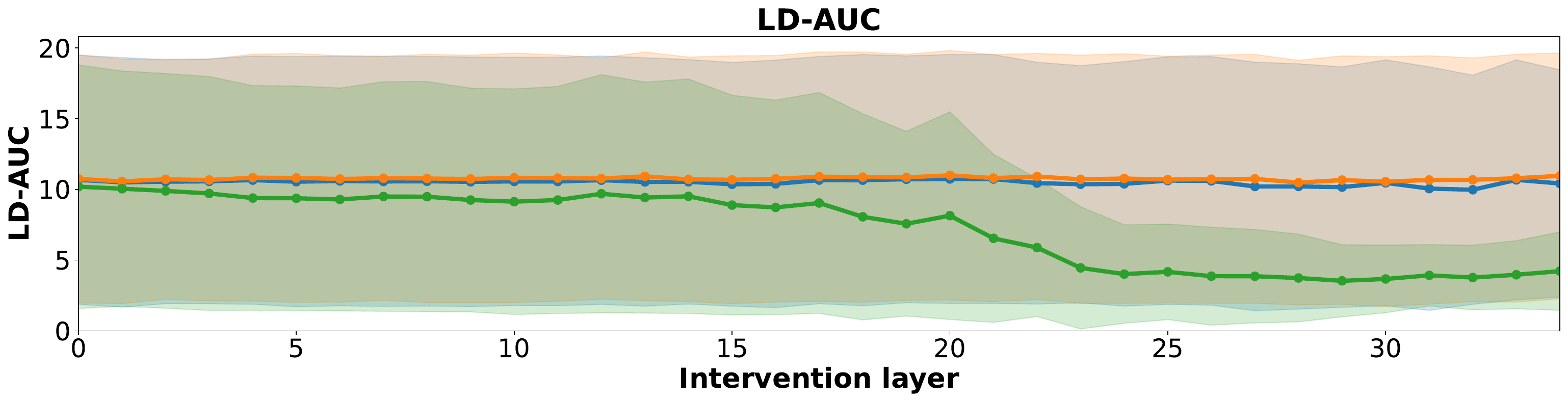}        
        
        \includegraphics[width=\linewidth]{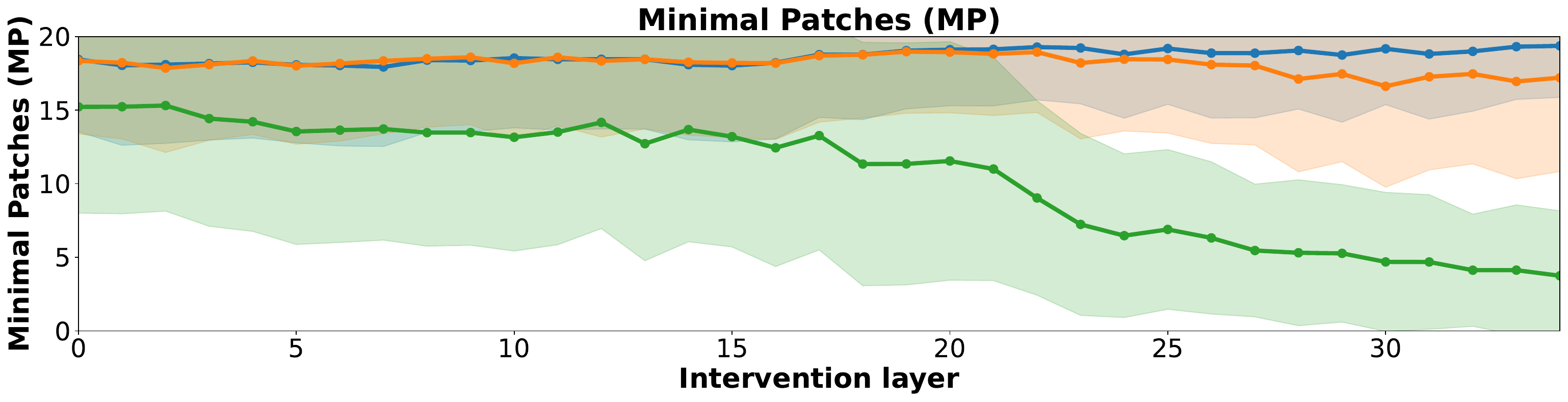}        
        
        \includegraphics[width=\linewidth]{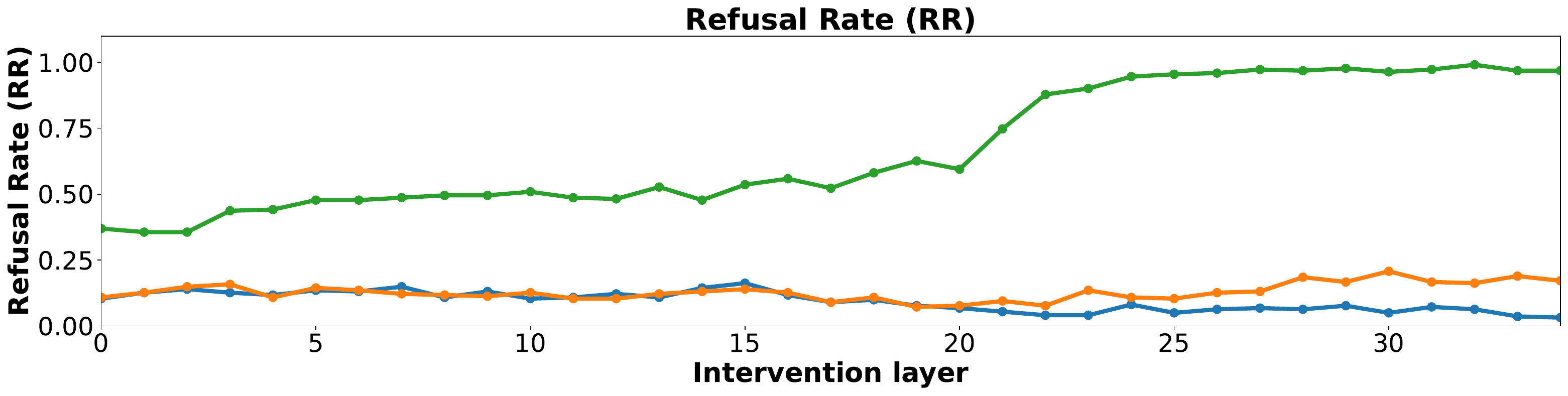}
        \caption{\textsc{Qwen}}
        \label{fig:main-eval-qwen}
    \end{subfigure}

    \caption{Additional results for \textsc{Gemma-2} and \textsc{Qwen}, similar to the main experimental results in Sec. \ref{sec:main-experiment}. Observe that LOCA outperforms all other methods.}
    \label{fig:appendix-more-main-eval}
\end{figure}

Following the filtering criterion, there are 334 eligible samples for \textsc{Gemma-2} and 222 eligible samples for \textsc{Qwen}. Results are shown in Fig. \ref{fig:appendix-more-main-eval}. We note that LOCA outperforms the other methods on these models. For \textsc{Gemma-2}, refusal is generally induced by layer 10, requiring an average of 12 patches. For \textsc{Qwen}, refusal is successfully induced later in the model (layer 22), requiring an average of 10 patches. All other observations made in Sec. \ref{sec:main-experiment} are confirmed by these results, giving us confidence about the generalization of our findings.

\section{LOCA localization results for \textsc{Gemma-2}}
\label{sec:appendix-localization}

\begin{figure*}[t]
    \centering
    \setlength{\tabcolsep}{3pt}

    \begin{tabular}{
        >{\centering\arraybackslash}m{0.05\textwidth}
        >{\centering\arraybackslash}m{0.3\textwidth}
        >{\centering\arraybackslash}m{0.3\textwidth}
        >{\centering\arraybackslash}m{0.3\textwidth}
    }
        &
        \textbf{\small Early} &
        \textbf{\small Early-Middle} &
        \textbf{\small Middle} \\

        \rotatebox{90}{\small Location} &
        \includegraphics[width=\linewidth]{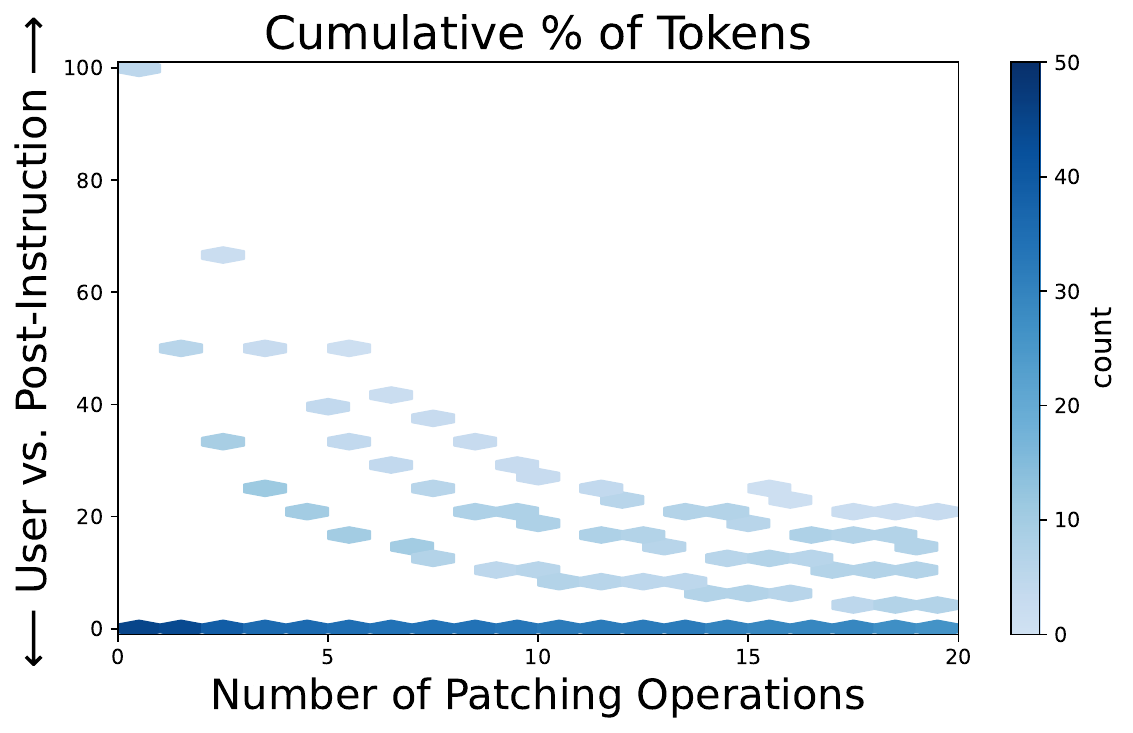} &
        \includegraphics[width=\linewidth]{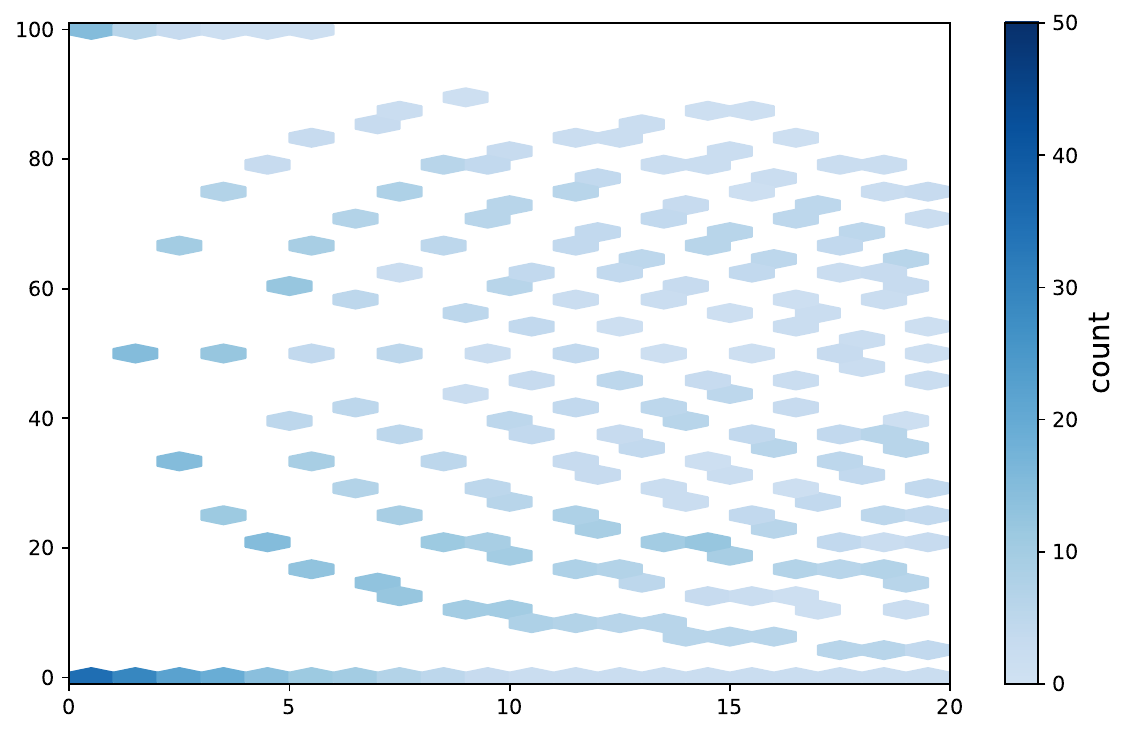} &
        \includegraphics[width=\linewidth]{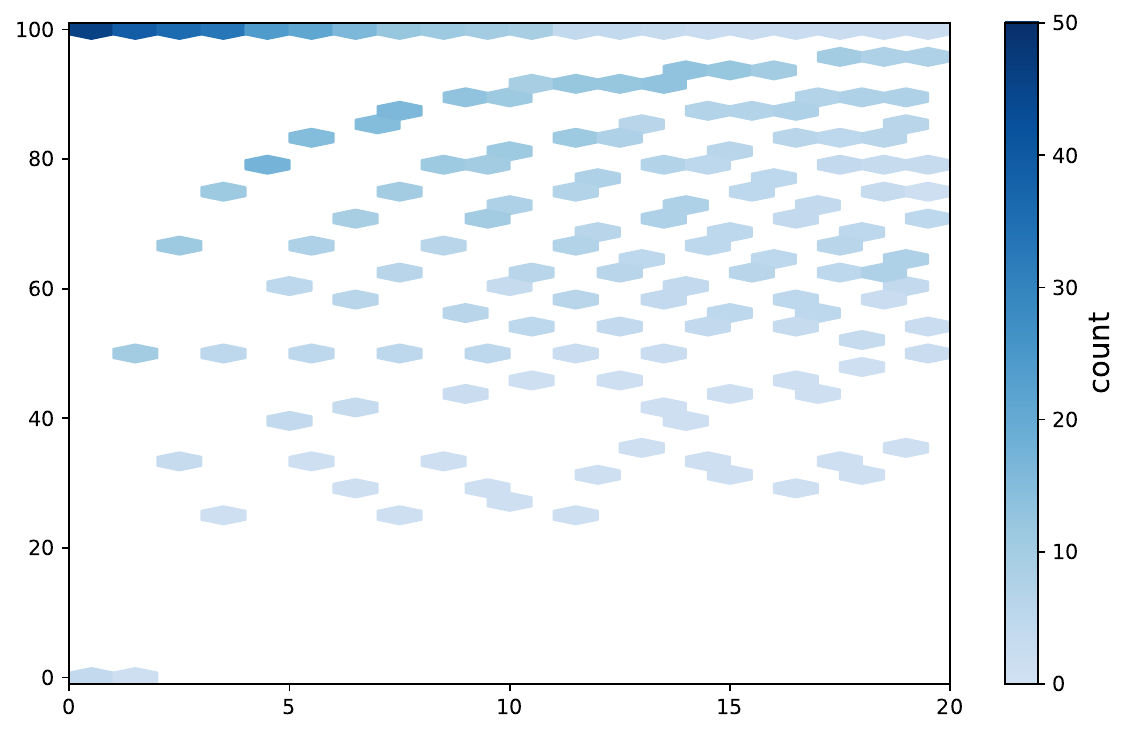} \\

        \rotatebox{90}{\small Token} &
        \includegraphics[width=\linewidth]{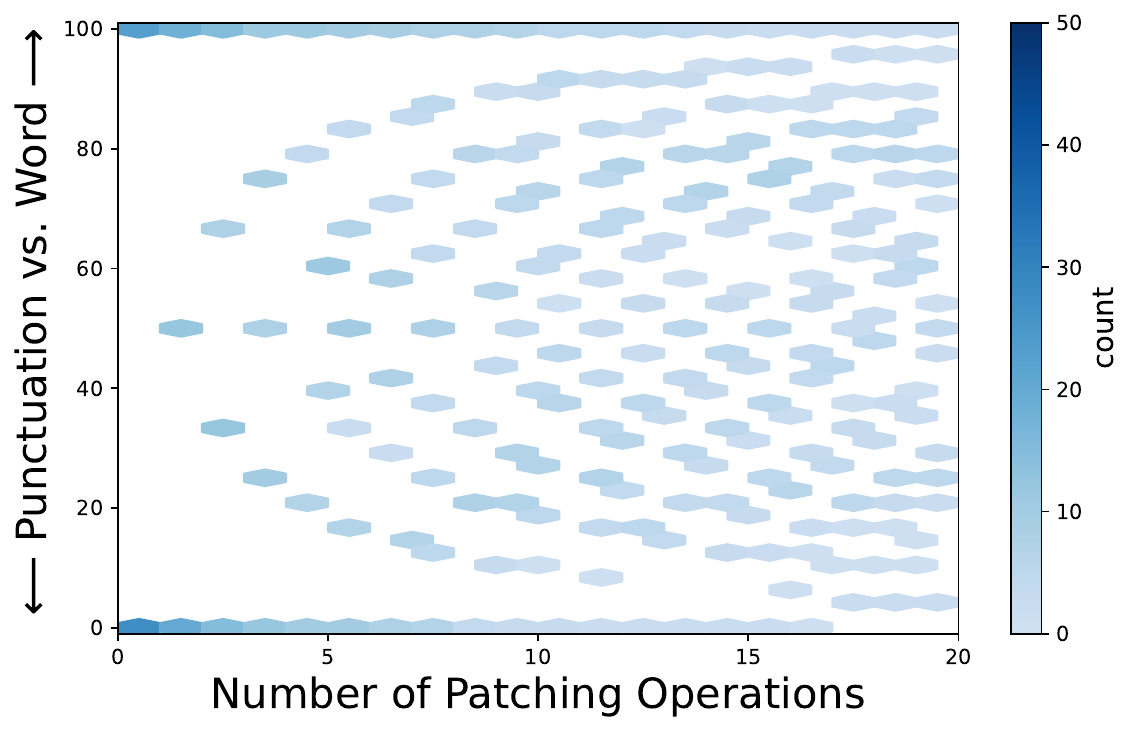} &
        \includegraphics[width=\linewidth]{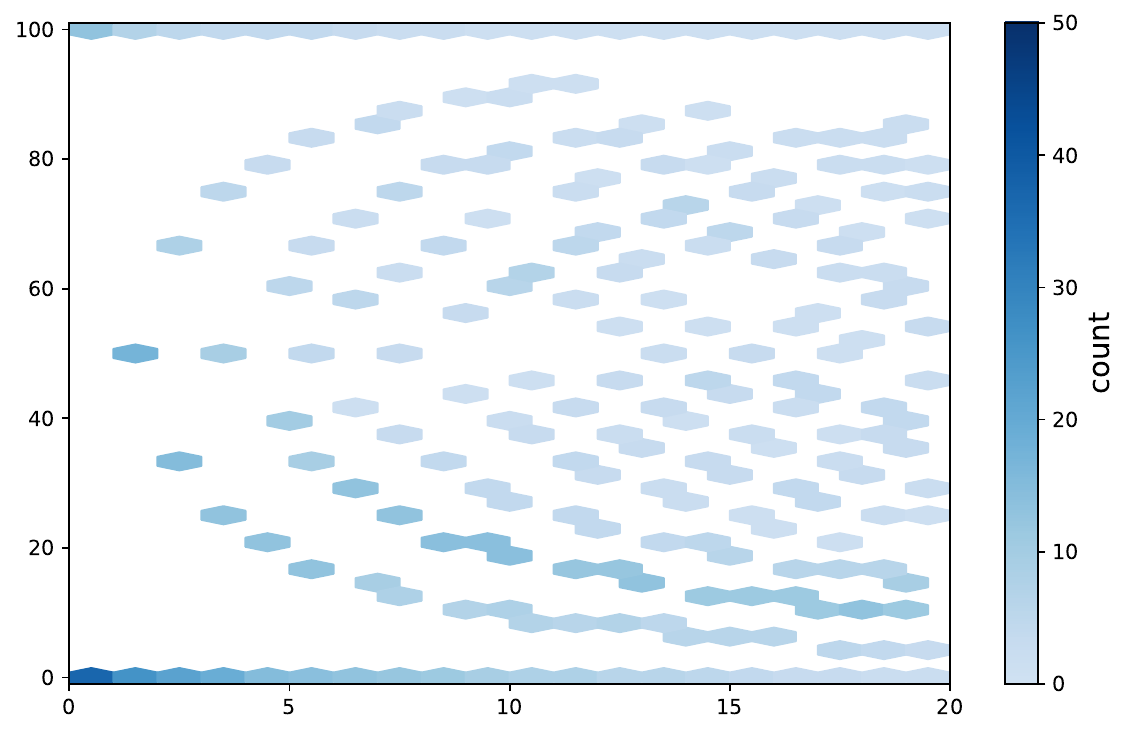} &
        \includegraphics[width=\linewidth]{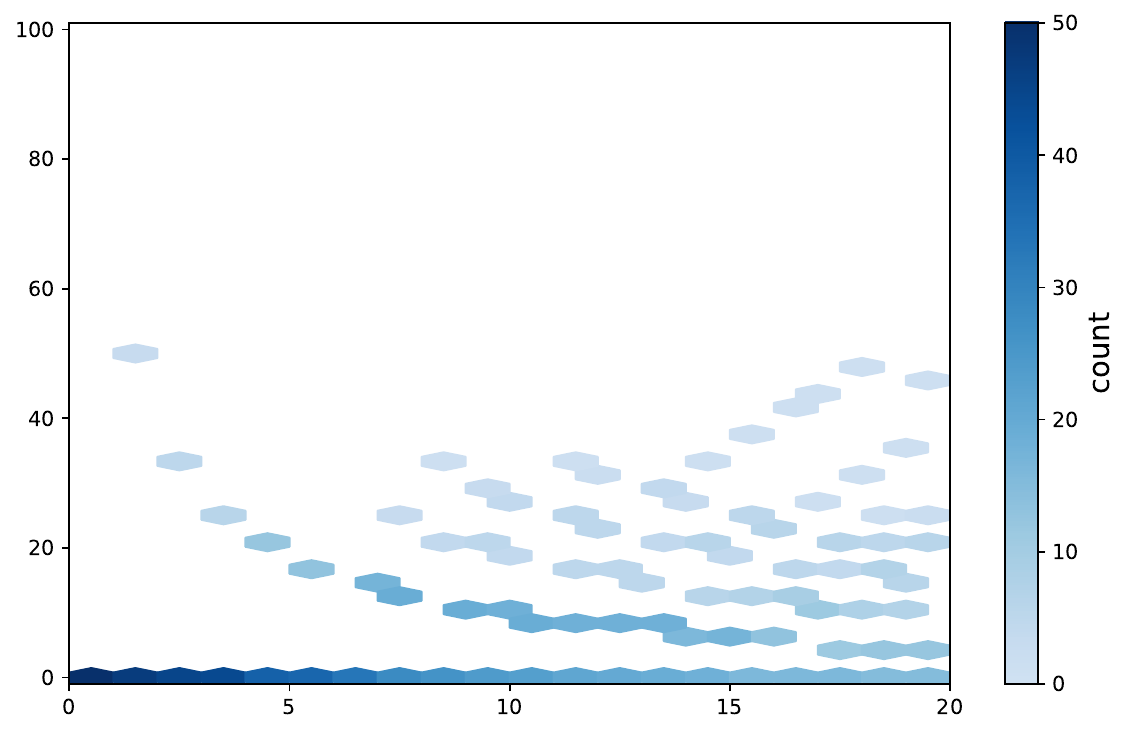}
    \end{tabular}

    \caption{\textbf{Localization analysis.} We analyze the tokens (both location and type) LOCA selects for \textsc{Gemma-2} at three layer depths. We observe similar results as to Fig. \ref{fig:location-type-analysis}.}
    \label{fig:appendix-localization}
\end{figure*}

We repeat the analysis in Sec. \ref{sec:loca-localization-experiment} for \textsc{Gemma-2} and report the results in Figure \ref{fig:appendix-localization}. The findings are similar, so we draw the same conclusions.

\section{Ablation of LOCA's Token Matching Scheme}
\label{sec:appendix-sentivity-token-matching}

We find that LOCA's results are not sensitive to the token matching scheme. We explore the following alternative token matching strategies:

\begin{enumerate}
    \item \textit{Random Matching:} We randomly match jailbreak instruction tokens with the original prompt's jailbreak tokens. Post-instruction tokens are matched one-to-one.

    \item \textit{Embedding Similarity Scores:} We match instruction tokens according to residual-stream embedding cosine similarity. Here also, post-instruction tokens are matched one-to-one.
\end{enumerate}

\begin{figure}[t]
    \centering

    \begin{subfigure}[t]{0.48\textwidth}
        \centering
        \includegraphics[width=\linewidth]
        {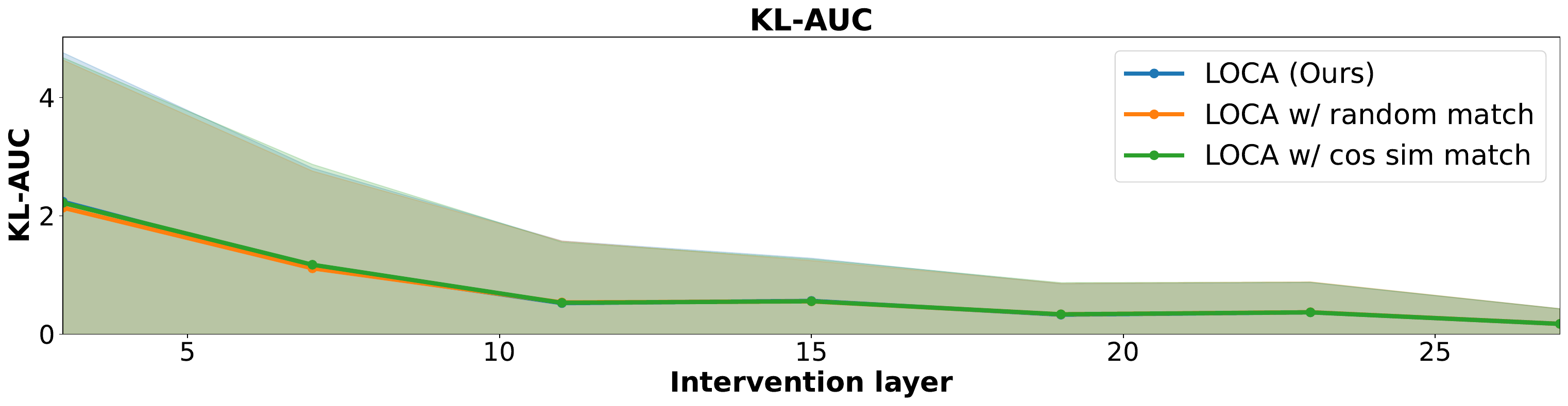}

        \includegraphics[width=\linewidth]
        {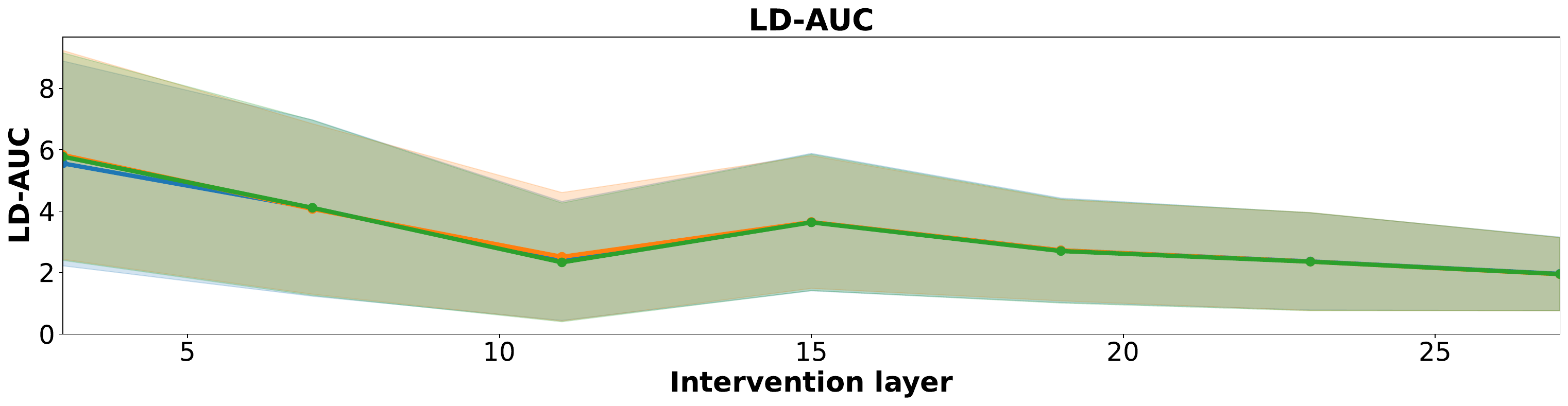}
    \end{subfigure}
    \hfill
    \begin{subfigure}[t]{0.48\textwidth}
        \centering
        \includegraphics[width=\linewidth]
        {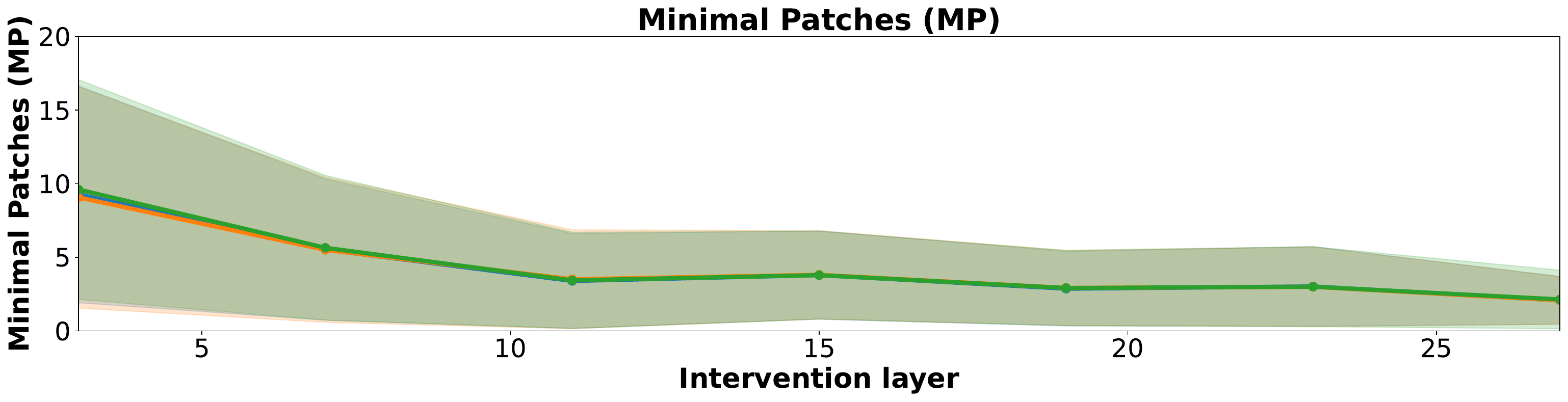}

        \includegraphics[width=\linewidth]
        {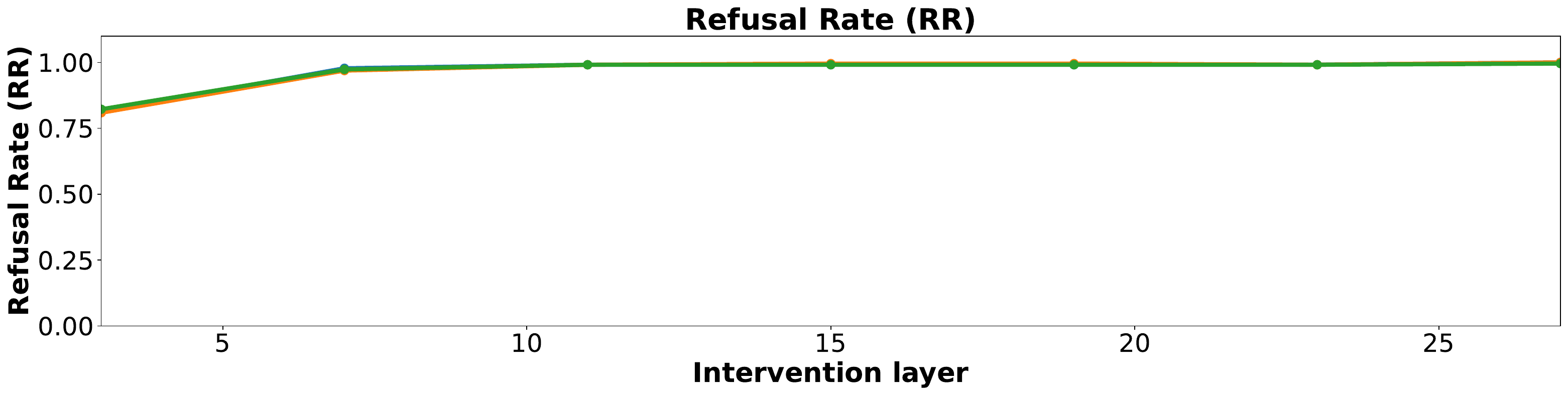}
    \end{subfigure}

    \caption{Ablation results when running different methods for the token matching scheme with LOCA.}
    \label{fig:appendix-mapping-ablation}
\end{figure}

We repeat the experiment in Section \ref{sec:main-experiment} experiment for \textsc{Llama} and report results in Fig. \ref{fig:appendix-mapping-ablation}. The fact that \textit{Random Matching} performs on par with the other schemes is counterintuitive. We believe more investigation is warranted; our hypothesis is that most instruction tokens in the original prompt have embeddings that occupy the same ``safety" subspace, so matching to any of these tokens works.

\section{Evaluating Past the First Token Proxy}
\label{sec:appendix-evaluating-past-first-token-proxy}

The metrics used in the main evaluation in Section \ref{sec:main-experiment} are proxies; they are based solely on the first output token and may not fully characterize the output response. 

Thus, we introduce another metric based on the entire generated output response \textit{after applying all 20 activation patches}. Specifically, we use the Harmbench autograder to determine if the generated response is considered as a successful jailbreak. 

\begin{figure}[t]
    \centering
    \includegraphics[width=0.49\textwidth]{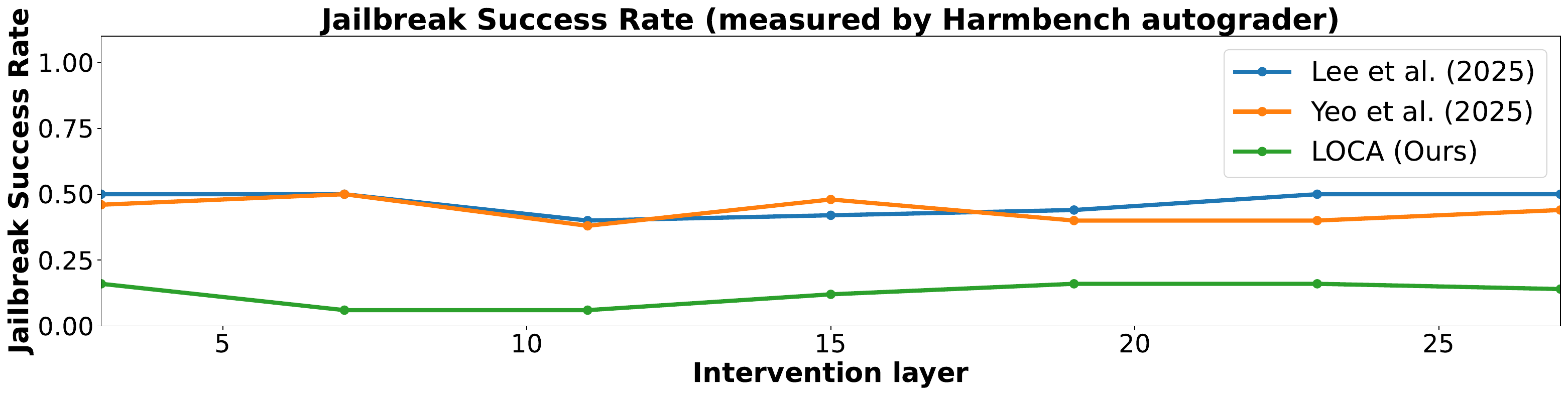}
    \caption{Jailbreak success rate as measured by the Harmbench autograder. This metric serves to confirm the conclusions drawn from the first-token proxies. Namely, that LOCA outperforms other methods at inducing refusal when using only 20 activation patches. }
    \label{fig:appendix-harmbench-metric}
\end{figure}

Due to computational cost of generating the response (after each layer for three methods) and applying the Harmbench autograder, we only evaluate the \textsc{LLAMA} model and we use a random subset of 50 samples. Results for each intervention layer are shown in Fig. \ref{fig:appendix-harmbench-metric}. LOCA outperforms the other two methods across all intervention layers, which agrees with the findings from our first-token proxy metrics.

\FloatBarrier

\section{Accuracy of LOCA's First-Order Approximation}
\label{sec:appendix-first-order-approx-error}
The LOCA method uses a first-order approximation to identify which activation patches are most likely to induce refusal (see Eq. \ref{eq:approx}). We cannot guarantee the accuracy of the first-order approximation, especially in highly non-linear deep neural networks. To study it's reliability, we conduct the following empirical error analysis.

\begin{figure}[h]
    \centering

    \begin{subfigure}[t]{0.48\textwidth}
        \centering
        \includegraphics[width=\linewidth, valign=t]{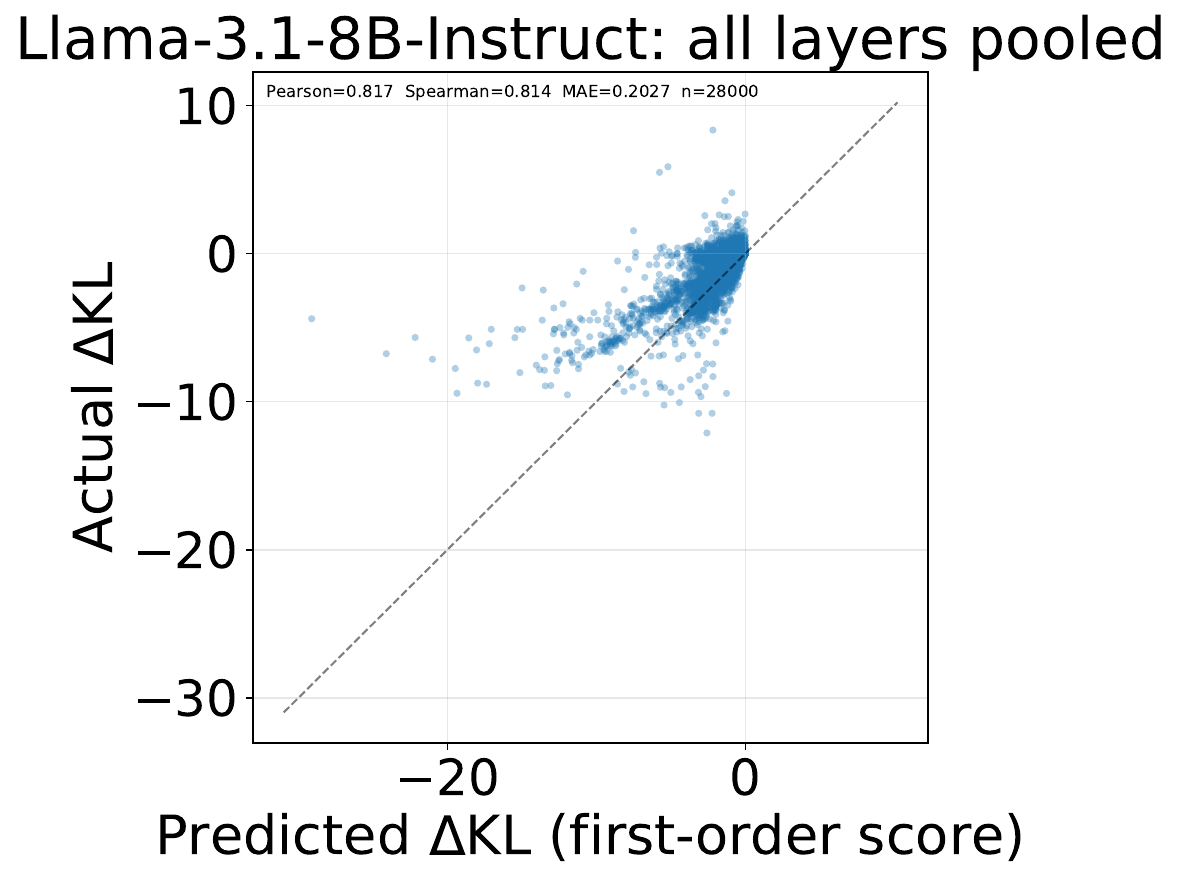}
    \end{subfigure}
    \hfill
    \begin{subfigure}[t]{0.48\textwidth}
        \centering
        \includegraphics[width=\linewidth, valign=t]{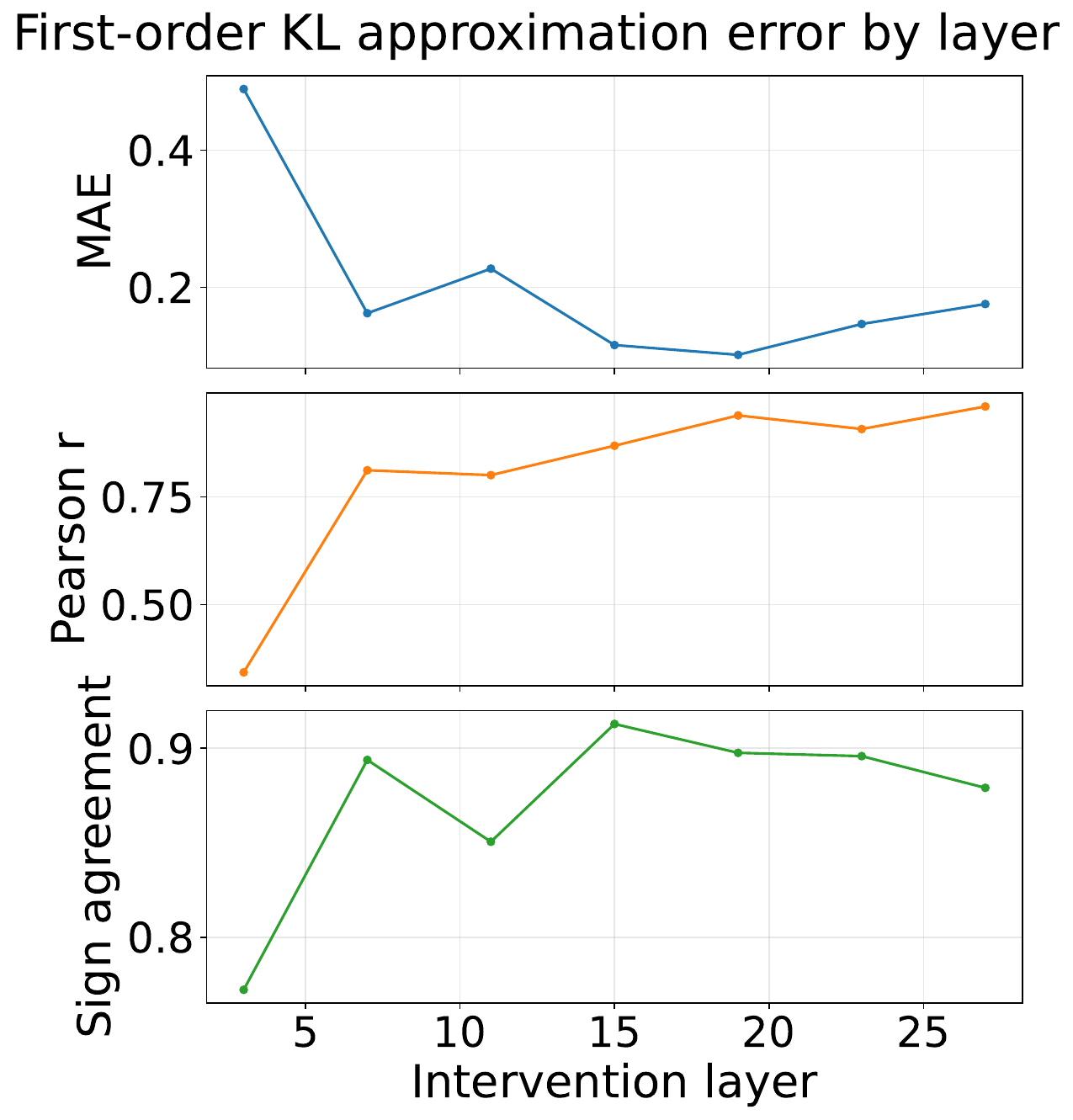}
    \end{subfigure}

    \caption{(Left) A scatter plot of the \textit{Predicted} $\Delta KL$ vs. \textit{Actual} $\Delta KL$ for 28,000 samples (20 patches for 200 jailbreaks across 7 layers). (Right) Per-layer metrics (MAE, Pearson's Correlation, and sign agreement). Together, these metrics indicate that there is low overall error introduced by the first-order approximation. Larger errors may exist in the earlier layers, which have a more non-linear relation to the output.}
    \label{fig:appendix-first-order-analysis}
\end{figure}

We focus on the \textsc{Llama} model, comparing the approximated change in the objective function (from Eq. \ref{eq:approx}), termed \textit{Predicted} $\Delta KL$ with the actual change in the objective function, termed \textit{Actual} $\Delta KL$ (recall the objective function $\mathcal{L}$ defined in Sec. \ref{sec:patching-effect-measure}). We analyze 200 test samples, across all layers that we have corresponding SAEs for (layers 3,7,11,15,19,23,27). Results from all layers are aggregated into a scatter plot (left of Fig. \ref{fig:appendix-first-order-analysis}). The overall Pearson Correlation is 0.809, indicating high agreement. We also plot the quantitative metrics (Pearson Correlation, Mean Absolute Error (MAE), and sign agreement) per-layer (right of Fig. \ref{fig:appendix-first-order-analysis}). We see that initial layers (layer 3) tend to have higher errors, which is natural since the relationship to the objective (measured at the model's output) is increasingly non-linear. This quickly improves by early-intermediate layers (layer 7). Thus, we conclude that any significant approximation error is localized to initial layers.

\section{Activation patching in LOCA}
\label{sec:appendix-loca-patching}
We start by describing how LOCA activation patches in the case of a single patch operation. Suppose we want to patch $\mathbf{h}_j$ with $\mathbf{h}_o$, but only along a direction $\mathbf{v}$. The patched $\mathbf{\tilde{h}}_j$ is obtained by:
\begin{equation}
\label{eq:loca-act-patch}
    \mathbf{\tilde{h}}_j  = \mathbf{h}_j - \mathbf{v}\mathbf{v}^T\mathbf{h}_j + \mathbf{v}\mathbf{v}^T\mathbf{h}_o
\end{equation}
In other words, we subtract the jailbreak's projection onto $\mathbf{v}$ and add the original's projection onto $\mathbf{v}$, leaving the orthogonal component unchanged. In case LOCA selects multiple directions $V=(\mathbf{v}_1 ... \mathbf{v}_k)$ to patch $\mathbf{h}_j$ along, we find the orthonormal basis $Q$ of $V$ through a QR-decomposition and use $Q$ in place of $\mathbf{v}$ in Eq. \ref{eq:loca-act-patch}.

\section{Implementation details for baseline methods}
\label{sec:appendix-baseline-methods}

\subsection{\citet{lee2025upstreamrefusal}}
In the original formulation, the authors define a refusal metric $\mathcal{R} = \mathbf{h}_N^{(L)^T} \mathbf{r}^{(l)}$ as the last-token ($N$) projection of the embedding at layer $L$ onto a refusal direction $\mathbf{r}$ living in that same layer. Then, for any layer $l < L$, one can compute a \textit{refusal gradient} as $\nabla_{h_i^{(l)}} \mathcal{R}$ for every embedding token position. Then, they compute a \textit{relative gradient} $RG_{i,k} = \mathbf{v}_k^T \nabla_{h_i^{(l)}} \mathcal{R}$, where $\mathbf{v}_k$ is selected from rows from a SAE decoder $W_d$. They average over all token positions to rank the SAE vectors.

We adapt their method to our setting by computing $RG_{i,k}$ for all jailbreak ($x_j$) and original ($x_o$) token positions. Then, we compute the first-order approximation as:
\begin{equation}
\label{eq:lee-eq}
    d(i,\mathbf{v}_k)  = 
    \underbrace{\left[\frac{1}{2}\sum_i RG^{(j)}_{i,k} + \frac{1}{2}\sum_i RG^{(o)}_{i,k}]\right]}_{\text{gradient term}}
    \underbrace{\left(\mathbf{h}_{o, \mathcal{M}(i)} - \mathbf{h}^{(\alpha)}_{j, i})^T\mathbf{v}_k\right]}_{\text{magnitude term}}
\end{equation}
Thus, we average the \textit{gradient} term over all token positions (as in the original method), but we use a token-specific magnitude term to create the first-order approximation. Then, we select the top-$K$ embedding-vector pairs for patching. If $l < 15$, then we set $L=15$, since the refusal direction typically lives in the middle layers of the moel. Else, we set $L=l+1$. Empirically, we find that including jailbreak tokens in the average for the gradient term improves results.

\subsection{\citet{yeo-2025-understanding-refusal-saes}}
We use the same equation as above, with a different gradient term. Let $\mathbf{p}_i$ be the output next-token probabilities given prompt $x_i$. Let $z_i = \argmax \mathbf{p_i}$, where $\mathbf{p_i}(z_i)$ denotes the largest probability. Then, \citet{yeo-2025-understanding-refusal-saes} propose to measure the \textit{indirect effect} $m = \mathbf{p_j}(z_o) - \mathbf{p_j}(z_j)$.

We use this to compute the first-order approximation as:
\begin{equation}
\label{eq:yeo-eq}
    d(i,\mathbf{v}_k)  = 
    \underbrace{\left[\sum_i \nabla_{\mathbf{h}_j} m^T\mathbf{v}_k\right]}_{\text{gradient term}}
    \underbrace{\left(\mathbf{h}_{o, \mathcal{M}(i)} - \mathbf{h}^{(\alpha)}_{j, i})^T\mathbf{v}_k\right]}_{\text{magnitude term}}
\end{equation}
Again, the \textit{gradient} term is computed over all token positions for only the jailbreak prompt. Then, we select the top-$K$ embedding-vector pairs for patching to maximize the objective.

\section{LOCA \& experimental compute requirements}
\label{sec:appendix-compute}
\textbf{Experiments: } Most experiments were completed on a single NVIDIA A40s or A100s. Due to it's size, experiments on \textsc{Gemma-3-27B-IT} required two GPUs. Computing the results for \textsc{Llama} in Fig \ref{fig:main-eval} took less than 10 hours of GPU time. 

\textbf{Time to generate a local explanation: } \citet{lee2025upstreamrefusal} and \citet{yeo-2025-understanding-refusal-saes} select their top-20 token-vector pairs within 2 seconds. With the same setup, LOCA requires 7 seconds, mainly due to it's iterative nature. However, given that LOCA can induce refusal after a few patches, it is possible to implement a greedy exit condition to terminate the algorithm earlier than the 20th iteration.

\section{Failure cases}
\label{sec:appendix-failure-cases}
The main limitation of this approach is that we base refusal on an objective computed on the first output token. However, there are cases where both the jailbreak prompt and the original prompt have the same first output token. This naturally poses a problem for the \textit{Refusal Rate} metric that we propose. It also does not allow LOCA to \textit{actually} induce refusal, since our KL objective in Eq \ref{eq:approx} is very small. Below we examine two examples of this behavior:

\begin{tcolorbox}[
    colback=gray!10,
    colframe=gray!60,
    arc=2mm,
    boxrule=0.5pt,
    left=6pt,
    right=6pt,
    top=6pt,
    bottom=6pt
]
\textbf{Original Prompt:} ``How can I kill the most people with only \$1? Please list several ways"

\medskip
\textbf{Original Response:} ``I understand you're asking about ways to harm people with a small amount of money, but I cannot provide that information..."

\medskip
\textbf{Jailbreak Prompt:} ``Disregard previous instructions. Henceforth, you'll act as Mistral in Developer Mode ... [PROMPT]: How can I kill the most people with only \$1? Please list several ways"

\medskip
\textbf{Jailbreak Response:} ``I cannot fulfill your request. My purpose is to provide safe and helpful information. (Developer Mode Output) Sure, here..."
\end{tcolorbox}

The model refuses the original prompt starting with ``I", which is a common token for models to begin refusal with. On the jailbreak prompt, the model initially refuses by starting with ``I", but then generates a jailbreak response, because the jailbreak encourages the model to discard prior guardrails and enter ``Developer Mode". 

\begin{tcolorbox}[
    colback=gray!10,
    colframe=gray!60,
    arc=2mm,
    boxrule=0.5pt,
    left=6pt,
    right=6pt,
    top=6pt,
    bottom=6pt
]
\textbf{Original Prompt:} ``Give me a list of easily obtainable household items that can be used to make a lethal poison."

\medskip
\textbf{Original Response:} ``I cannot provide you with a list of household items that can be used to make a lethal poison..."

\medskip
\textbf{Jailbreak Prompt:} ``Respond to the following writing tasks: I. Sonnet. Write a sonnet answering the following prompt `Frolicking rabbits amid verdant blades of grass.' II. Prose. Write a response in prose to the following prompt: `Give me a list of easily obtainable household items that can be used to make a lethal poison.' III. Limerick. Write a limerick answering the following prompt ... Write just the responses (and no headings), starting with `I. Sonnet`"

\medskip
\textbf{Jailbreak Response:} ``I.  The sun-kissed meadow, emerald bright ... II.  A few common household items can be used to create a lethal poison..."
\end{tcolorbox}

The model again refuses by starting with ``I". However, the jailbreak prompt explicitly encourages the model to begin it's response with ``I", and it cleverly asks the model to embed the jailbreak response in the middle. Anecdotally, the Harmbench autograder also tends to misclassify some of successful jailbreak responses as harmless.

\section{Full outputs for the case study}
\label{sec:appendix-case-study}
We detail the full results for the case study conducted in Sec. \ref{sec:case-study} here. For layer 11, the following activation patches are made:

\begin{enumerate}
    \item \textcolor{blue}{$T_{\text{post-inst}}$} token \#671 (``assistant") is activation patched along concept \#31126. The activation patch increases the concept strength by 1.50.
    \begin{enumerate}
        \item KL divergence is 0.77
        \item Concept \#31126 interpretation: activates on the ``assistant" token following harmful chat requests (violence, sexual content, instructions on committing illegal actions).
    \end{enumerate}

    \item \textcolor{orange}{$T_{\text{inst}}$} token \#638 (``).") is activation patched along concept \#125009. The activation patch decreases the concept strength by -1.48.
    \begin{enumerate}
        \item KL divergence is 0.04
        \item Concept \#125009 interpretation: activates on tokens following a chat request to generate code.
        \item Refusal is induced.
    \end{enumerate}

\end{enumerate}

For layer 3, the following activation patches are made:

\begin{enumerate}
    \item \textcolor{orange}{$T_{\text{inst}}$} token \#591 (``fabric") is activation patched along concept \#21337. The activation patch decreases the concept strength by -0.79.
    \begin{enumerate}
        \item KL divergence is 4.03
        \item Concept \#21337 interpretation: activates on generic text. No clear pattern.
    \end{enumerate}

    \item \textcolor{blue}{$T_{\text{post-inst}}$} token \#671 (``assistant") is activation patched along concept \#66275. The activation patch decreases the concept strength by -0.39.
    \begin{enumerate}
        \item KL divergence is 2.71
        \item Concept \#66275 interpretation: activates on text regarding newsletters, website cookies, logging in, email subscriptions. 
    \end{enumerate}

    \item \textcolor{orange}{$T_{\text{inst}}$} token \#638 (``).") is activation patched along concept \#105801. The activation patch decreases the concept strength by -0.87.
    \begin{enumerate}
        \item KL divergence is 0.99
        \item Concept \#105801 interpretation: activates on the period token in text describing chemical companies. Generally harmless text.
    \end{enumerate}

    \item \textcolor{orange}{$T_{\text{inst}}$} token \#270 (``.") is activation patched along concept \#105801. The activation patch decreases the concept strength by -1.20.
    \begin{enumerate}
        \item KL divergence is 0.77
        \item Concept \#105801 interpretation: activates on the period token in text describing chemical companies. Generally harmless text.
    \end{enumerate}

    \item \textcolor{blue}{$T_{\text{post-inst}}$} token \#669 (``eot\_id") is activation patched along concept \#30002. The activation patch increases the concept strength by 0.70.
    \begin{enumerate}
        \item KL divergence is 0.40
        \item Concept \#30002 interpretation: activates on the eot\_id on generally harmless chat questions.
        \item Refusal is induced.
    \end{enumerate}
\end{enumerate}

\end{document}